%% file: samplepaper.tex
\documentclass[runningheads]{llncs}
\usepackage{graphicx}
\usepackage{tikz}
\usetikzlibrary{shapes,arrows,positioning,decorations.markings,fit}
\usepackage{amsmath}
\usepackage{amsfonts}
\usepackage{amssymb}
\usepackage{tabularx}
\usepackage{multirow}
\usepackage{caption}
\usepackage{subcaption}
\usepackage{todonotes}
\usepackage{multicol}
\usepackage{dsfont}
\usepackage{booktabs}

\newcommand{\mat}[1]{{\boldsymbol{{#1}}}} 
\begin{document}
\title{A Recurrent Neural Network Survival Model: \\ Predicting Web User Return Time}
\titlerunning{A Recurrent Neural Network Survival Model}
%
\author{Georg L. Grob\inst{1} \and
\^{A}ngelo Cardoso\inst{2}\thanks{Now with Vodafone Research and ISR, IST, Universidade de Lisboa} \and
C. H. Bryan Liu\inst{2} \and
Duncan A. Little\inst{2} \and
Benjamin Paul Chamberlain\inst{1,2}
}

\authorrunning{G. L. Grob et al.}
%
\institute{Imperial College London, London, UK \hspace*{10pt} \email{grobgl@gmail.com}\and
ASOS.com, London, UK \hspace*{10pt}
\email{bryan.liu@asos.com}}
\maketitle              
\begin{abstract}
The size of a website's active user base directly affects its value.
Thus, it is important to monitor and influence a user's likelihood to return to a site. Essential to this is predicting \textit{when} a user will return.
Current state of the art approaches to solve this problem come in two flavors: (1) Recurrent Neural Network (RNN) based solutions and (2) survival analysis methods. We observe that both techniques are severely limited when applied to this problem. Survival models can only incorporate aggregate representations of users instead of automatically learning a representation directly from a raw time series of user actions. RNNs can automatically learn features, but can not be directly trained with examples of non-returning users who have no target value for their return time.
We develop a novel RNN survival model that removes the limitations of the state of the art methods. We demonstrate that this model can successfully be applied to return time prediction on a large e-commerce dataset with a superior ability to discriminate between returning and non-returning users than either method applied in isolation.

\end{abstract}

\section{Introduction}

Successful websites must understand the needs, preferences and characteristics of their users. A key characteristic is \textit{if} and \textit{when} a user will return. 
Predicting user return time allows a business to put in place measures to minimize absences and maximize per user return probabilities. Techniques to do this include timely incentives, personalized experiences~\cite{Du:2015:TRR:2969442.2969629} and rapid identification that a user is losing interest in a service~\cite{Benson:2016:MUC:2872427.2883024}. A related problem is user churn prediction for non-contractual services. In this case a return time threshold can be set, over which a user is deemed to have churned. Similar prevention measures are often put in place for users with high churn risks~\cite{Chamberlain:2017:CLV:3097983.3098123}. 

This paper focuses on predicting user return time for a website based on a time series of user sessions. The sessions have additional features and so form a \textit{marked temporal point processes} for each user. As some users do not return within the measurement period, their return times are regarded as \textit{censored}. The presence of missing labels makes the application of standard supervised machine learning techniques difficult. However, in the field of survival analysis, an elegant solution to the missing label problem exists which has been transferred to a variety of settings~\cite{ishwaran2008}~\cite{deepSurvivalAnalysis}.

Recurrent Neural Networks (RNNs) have achieved significant advances in sequence modelling, achieving state-of-art performance in a number of tasks \cite{sutskever2014sequence,vinyals2015show,graves2009novel}. Much of the power of RNNs lie in their ability to automatically extract high order temporal features from sequences of information. Many complex temporal patterns of web user behaviour can exist. This can include noisy oscillations of activity with periods of weeks, months, years, pay days, festivals and many more beside. Exhaustively handcrafting exotic temporal features is very challenging and so it is highly desirable to employ a method that can automatically learn temporal features.

In this paper, we predict user return time by constructing a recurrent neural network-based survival model. This model combines useful aspects of both RNNs and survival analysis models. RNNs automatically learn high order temporal features from user `sessions' and their associated features. They can not however, be trained with examples of users who do not return or the time since a user's latest session. Survival models can include information on users who do not return (right-censored users) and the time since their last session, but cannot learn from a sequence of events.

Our main contribution is to develop a RNN-based survival model which incorporates the advantages of using a RNN and of using survival analysis. The model is trained on sequences of sessions and can also be trained with examples of non-returning users. We show that this combined model outperforms both RNNs and survival analysis employed in isolation. 
We also provide the code implementation for use by the wider research community.\footnote{\texttt{\url{https://github.com/grobgl/rnnsm}}}


\section{Background}

\begin{figure}
\centering
\begin{tikzpicture}[thick, scale=0.8, every node/.style={transform shape}]
\node[draw, minimum width=15em, minimum height=4em, text width=15em, fill=green!10] (obs) {Observation\\ window};
\node[draw, left=0em of obs.south east, anchor=south east, minimum width=4em, minimum height=3em, text width=6em, fill=green!40] (act) {Activity\\ window};
\node[draw, right=-0.4pt of obs.east, minimum width=4em, minimum height=4em, text width=6em, fill=blue!10] (pred) {Prediction\\ window};
\draw[->,thick] ([yshift=-1em, xshift=-2em]obs.south west) -- ([xshift=2em, yshift=-1em]pred.south east);
\draw[thick] ([yshift=-1.5em]obs.south west) -- node[label=below:{0}] {} ([yshift=-.5em]obs.south west);
\draw[thick] ([yshift=-1.5em]act.south west) -- node[label=below:{$t_a$}] {} ([yshift=-.5em]act.south west);
\draw[thick] ([yshift=-1.5em]pred.south west) -- node[label=below:{$t_p$}] {} ([yshift=-.5em]pred.south west);
\draw[thick] ([yshift=-1.5em]pred.south east) -- node[label=below:{$t_n$}] {} ([yshift=-.5em]pred.south east);
\end{tikzpicture}
\vspace*{-10pt}
\caption{Illustration of the observation window, the activity window, and the prediction window. The x-axis denotes time. We use observations occurring within the observation window $\left[0,t_p\right]$ on users active in the activity window $\left[t_a, t_p\right]$. We then predict their return times with respect to the prediction window $\left(t_p,t_n\right]$.}
\vspace*{-5pt}
\label{fig:windows}
\end{figure}
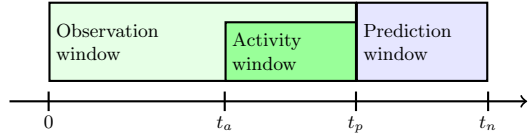

We are interested in predicting the return times of users to a website. We select a period of time during which users must be observed visiting a site and call this the \textit{Activity window}. We declare a separate disjoint period of time called the \textit{Prediction window} from which we generate return time labels and make predictions and both windows are illustrated in Figure~\ref{fig:windows}. There are necessarily two types of user: returning and non-returning. We consider a user as non-returning if they do not have any sessions within the Prediction window, and a returning user as those who do. As suggested by Wangperawong et al.~\cite{1604.05377} we record data for some time preceding the Activity window (called the \textit{Observation window}) to avoid producing a model that would predominantly predict non-returning users. This set up allows us to define the set of users active in the activity window; $\mathcal{C}$, the set of returning users, $\mathcal{C}_{\textrm{ret}}$ and the set of non-returning users $\mathcal{C}_{\textrm{non-ret}}$.

We follow the definition of return time suggested by Kapoor et al.~\cite{Kapoor:2014:HBA:2623330.2623348}, that is, $d_{j+1}^{\left(i\right)} = t_{j+1}^{\left(i\right)} - t_{j}^{\left(i\right)}$, denotes the period between the end of the $i$th user's $j$th session and the beginning of the succeeding session. A session occurs when a user browses a website or mobile app. The $i$th user's $j$th session, $s_j^{\left(i\right)} = \left(t_j^{\left(i\right)}, \vec y_j^{\left(i\right)}\right)$, has an associated start time $t_j^{\left(i\right)} \in \left[0,\infty\right)$ and a vector of $n$ features $\vec y_j^{\left(i\right)}$. A user's browsing history can therefore be represented as a sequence of sessions, $\mathcal{S}^{\left(i\right)}$, where $\mathcal{S}^{\left(i\right)} = \left(s_1^{\left(i\right)}, s_2^{\left(i\right)}, \ldots\right)$.

\subsection{Survival analysis}\label{sec:survival_analysis}
Survival analysis models the time until an event of interest occurs \cite{klein2005survival}. In the context of users' return time prediction, return time is equivalent to survival time.

\paragraph{Harzard and survival functions}
Here we clarify the notation and standard results used throughout the paper, as defined in~\cite{Rodriguez2010SurvivalModels}. $T$ is a random variable denoting the lifetime of an individual. The probability density function  for $T$, corresponding to the probability of the event of interest occurring at time $t$ is written as:
\begin{align}\label{eq:pdf}
f\left(t\right) =\lim_{\delta t\rightarrow 0} \frac{P\left(t < T \leq t + \delta t\right)}{\delta t},
\end{align}
with $F\left(t\right)$ as the corresponding cumulative density function.
The survival function $S\left(t\right)$, denoting the probability of the event of interest not having occurred by time $t$, is defined as:
\begin{align}
S\left(t\right) = P\left(T\geq t\right) = 1 - F\left(t\right) = \int_t^\infty f\left(z\right) \,\textrm{d}z .
\label{eq:survival_function}
\end{align}
The hazard function, which models the instantaneous rate of occurrence given that the event of interest did not occur until time~$t$, is defined as:
\begin{align}
\lambda(t) = \lim_{\delta t\rightarrow 0}\frac{P\left(t\le T< t + \delta t\middle|T\ge t\right)}{\delta t} 
= \frac{f(t)}{1-F(t)} = \frac{f(t)}{S(t)} .
\label{eq:hazard_function}
\end{align}

The hazard function is related to the survival function by
\begin{align}
\label{eq:survival_to_hazard}
-\frac{d\log S(t)}{dt} = \lambda(t) .
\end{align}

\paragraph{Censoring}\label{sec:censoring}
Censoring occurs when labels are partially observed. Klein and Moeschberger~\cite{klein2005survival} provide two definitions of censoring (1) an \textit{uncensored observation} when the label value is observed and (2) \textit{right-censored observation} when the label value is only known to be above an observed value.

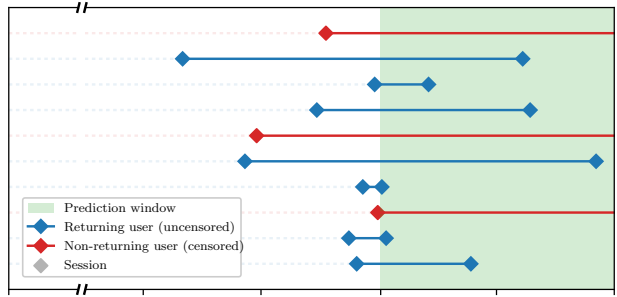
\begin{figure}
\vspace*{-5pt}
    \centering
    \noindent\resizebox{0.75\columnwidth}{!}{
        \hspace{-6pt}
        \input{censorship.pgf}}
        \vspace{-15pt}
        \caption{Visualisation of censored return times. The horizontal axis represents time and the vertical axis different users. The shaded area represents the prediction window. The solid lines represent the return time after users' last sessions in the observation time frame -- the value we are aiming to predict. The return times of users that do not return in the prediction time frame are censored. We do not know their actual return time. However, we do know that their return time spans at least across the entire prediction window.}
\label{fig:censorship}
\end{figure}

In the context of return time prediction, some users do not return to the website during the observation period. We label these users as non-returning, but some will return to the website after the observation period. Figure~\ref{fig:censorship} shows a selection of returning and non-returning users. To estimate the average return time, it is not sufficient to only include returning users as this underestimates the value for all users. Including non-returning users' time since their latest session still underestimates the true average return time.

To address this problem, we must incorporate censoring into our survival model. This is achieved using a likelihood function that has separate terms to account for censoring:
\begin{align}
\label{eq:survival_likelihood}
\begin{split}
L\left(\vec\theta\right) & =
  \prod_{i \in \text{unc.}}\hspace{-.5em} P\left(T=T_i\middle|\vec\theta\right)\,
  \prod_{j \in \text{r.c.}}\hspace{-.3em} P\left(T>T_j\middle|\vec\theta\right)\,
    = \prod_{i \in \text{unc.}}\hspace{-.5em} f\left(T_i\middle|\vec\theta\right) \,
      \prod_{j \in \text{r.c.}}\hspace{-.3em} S\left(T_j\middle|\vec\theta\right),
\end{split}
\end{align}
where $\vec\theta$ is a vector of model parameters. \textit{unc.} and \textit{r.c.} denote the uncensored and right-censored observations respectively. $T_i$ and $T_j$ denote the exact value of the uncensored observation and the minimum value of the right-censored observation respectively. For simplicity we assume there are no observations which are subject to other types of censoring.

\subsection{Cox Proportional Hazards Model}\label{sec:cox_ph_model}
The Cox proportional hazards model \cite{10.2307/2985181} is a popular survival regression model. It is applied by Kapoor et al.~\cite{Kapoor:2014:HBA:2623330.2623348} to predict the return time of users.

The model assumes that one or more given covariates have (different) multiplicative effects on a base hazard. The model can be defined in terms of its hazard function:
\begin{align}
    \lambda\left(t\middle|\vec x\right) = \lambda_0(t) \exp\left(\vec x^T\vec \beta\right) ,
\end{align}

where $\lambda_0\left(t\right)$ is the baseline hazard, $\vec x$ a vector of covariates, and $\beta_i$ the multiplier for covariate $x_i$. The model implicitly assumes the parameters for each covariate can be estimated without considering the baseline hazard function.

Various methods to estimate the multipliers~$\vec{\beta}$ exist \cite{cox75,10.2307/2334538,breslow1974covariance}, and we use that featured in the \textit{lifelines} library \cite{Davidson-Pilon2016Lifelines}, which maximises the Efron's partial likelihood \cite{efron1977efficiency} for $\vec{\beta}$ to obtain the estimate~$\hat{\vec\beta}$.

The estimated baseline hazard is then computed as described by Cox and Oakes \cite{cox1984analysis}:
\begin{align}
\hat \lambda_0\left(t_{\left(i\right)}\right) = \frac{d_{\left(i\right)}}{\sum_{j \in \mathcal{R}\left(t_{\left(i\right)}\right)} \exp\left(\hat{\vec\beta}^T \vec x_j\right)} ,
\end{align}
where~$t_{\left(i\right)}$ denotes the $i$ unique ordered time of an event of interest, $d_{\left(i\right)}$ is the number of events of interest occurring at time~$t_{\left(i\right)}$, and~$\mathcal{R}\left(t_{\left(i\right)}\right)$ is the set of individuals for whom the event of interest has not occurred by time $t_{\left(i\right)}$.
The users' return times can then be estimated by calculating their expected survival time.

The Cox proportional hazards model is particularly suitable for this problem as it allow us to include right-censored observations (users who appeared not returning) as training examples. Their value is only known to be above a certain value, which corresponds to the time between the last session in the observation time frame and the end of the prediction time frame.

\subsection{Recurrent Neural Networks, LSTM, and Embeddings}
A recurrent neural network (RNN) is a feedforward neural network where the output of a hidden unit at the current timestep is fed back into the hidden unit so that it forms part of the input for  the preceding timesteps. This allows RNNs to learn from sequences of events ordered chronologically or otherwise. The power of an RNN lies in its ability to learn from the current state of the sequence within the context of what has gone before. This context is stored as an internal memory within the hidden units of the RNN. For modelling time series data, sequences of events are discretised in time \cite{han2004prediction,cai2007time,chandra2012cooperative}.

Long Short-Term Memory (LSTM) units \cite{doi:10.1162/neco.1997.9.8.1735} 
and Gated Recurrent Units (GRUs) \cite{cho2014properties} were developed to overcome the problems associated with learning long-term dependencies in traditional RNNs \cite{Bengio:1994:LLD:2325857.2328340}. LSTMs and GRUs solve this issue by learning what information they should keep from the previous time step and what information they should forget.

It is also common to add embedding layer(s) to neural networks to transform large and sparse information info dense representations before the actual training of the network \cite{covington45530,flunkert2017deepar}. The embedding layer will automatically learn features (in the form of the dense representation's individual components) for the neural network's consumption.

First popularised by Mikolov et al.~\cite{mikolov2013efficient} to encode words in documents, embedding layers have been shown to encode various categorical features with a large number of possible values well~\cite{Li:2017:NNC:3041021.3054200,cheng2016wide}. In this paper we use the embedding layer implementation in Keras~\cite{chollet2015keras} with a TensorFlow backend.

\subsection{Recurrent Temporal Point Processes}

Temporal point processes model times of reoccurring events, which may have markers (features) associated with them, such as click rates and duration for web sessions. Manzoor et al. \cite{Manzoor:2017:RTT:3097983.3098104} modeled both the timing and the category of a user's next purchase given a history of such events using a Hawkes process, which assumes the occurrence of past events increases the likelihood of future occurrences~\cite{10.2307/2334319}.

Du et al. \cite{Du2016} propose the recurrent marked temporal point process (RMTPP) to predict both timings and markers (non-aggregated features) of future events given a history of such events. They assume events have exactly one discrete marker and employ RNNs to find a representation for the event history, which then serves as input to the hazard function. The paper demonstrates that such process can be applied in a wide variety of settings, including return times of users of a music website.

The RMTPP model is formulated as follows. Let $\mathcal{H}_t$ be the history of events up to time $t$, containing event pairs $\left(t_j, y_j\right)_{j\in \mathbb{Z}^+}$ denoting the event timing and marker respectively. The conditional density function corresponds to the likelihood of an event of type $y$ happening at time $t$: $f^*\left(t,y\right) = f\left(t,y\middle|\mathcal{H}_t\right)$.\footnote{The $*$-notation is used to denote the conditioning on the history.}

A compact representation $\vec h_j$ of the history up to the $j$th event is found through processing a sequence of events $\mathcal{S}=\left(t_j, y_j\right)_{j=1}^n$ with an RNN. This allows the representation of the conditional density of the next event time as:
\begin{align}
f^*\left(t_{j+1}\right) = f\left(t_{j+1}\middle|\vec h_j\right) .
\end{align}

Given $\vec h_j$, the hazard function of the RMTPP is defined as follow:
\begin{align}\label{align:rmtpp_hazard}
\lambda^*\left(t\right) = \exp \Big(\underbrace{\vec v^{\left( t\right)\top} \vec h_j}_{\substack{\text{past} \\ \text{influence}}} + \underbrace{w \left(t-t_j\right)}_{\substack{\text{current} \\ \text{influence}}} + \underbrace{b^{\left(t\right)}}_{\substack{\text{base} \\ \text{intensity}}}\Big) ,
\end{align}
where~$\vec v^{\left(t\right)}$ is the hidden representation in the recurrent layer in the RNN (which takes only~$\vec h_j$, the representation of past history, into account), $t - t_j$ is the absence time at the time of prediction (the current information), $w$ is a specified weight balancing the influence from the past history from that of the current information,\footnote{Du et al.~\cite{Du2016} specified a different fixed value for $w$ in their models fitted under different datasets. We use Bayesian Optimisation to find the best $w$ in our experiments.} and~$b^{\left(t\right)}$ is the base intensity (or bias) term of the recurrent layer.

\tikzstyle{box_layer} = [draw, fill=gray!20, thick, minimum width=4.5em, node distance=3em, outer sep=.1cm, inner sep=1em, rounded corners]
\tikzstyle{loss_layer} = [draw, thick, minimum width=3.5em, node distance=2cm, outer sep=.1cm, inner sep=.5em, rounded corners]
\tikzstyle{small_box_layer} = [draw, fill=gray!20, thick, minimum width=1.5em, node distance=3em, outer sep=.1cm, inner sep=.5em, rounded corners]

The conditional density is given by swapping the terms in Equation~\eqref{eq:hazard_function} and integrating Equation~\eqref{eq:survival_to_hazard}:
\begin{align}
f^*\left(t\right)
& = \lambda^* \left(t\right) \exp \left(-\int_{t_j}^t \lambda^*\left(\tau\right) d\tau\right) \nonumber \\
& = \exp \bigg(\vec v^{\left(t\right)\top} \vec h_j + w\left(t-t_j\right) + b^{\left(t\right)} \nonumber 
\; + \frac{1}{w}\exp \left(\vec v^{\left(t\right)\top} \vec h_j + b^{\left(t\right)}\right) \nonumber \\ 
&\qquad\quad - \frac{1}{w} \exp\left(\vec v^{\left(t\right)\top} \vec h_j + w\left(t - t_j\right) + b^{\left(t\right)}\right)\bigg) . 
\label{align:rmtpp_time_likelihood}
\end{align}

The timings of the next event can then be estimated by taking the expectation of the conditional density function:
\begin{align}
\hat t_{j+1} = \int_{t_j}^\infty t f^*\left(t\right) \textrm{d}t .
\end{align}

\begin{figure}
\vspace*{-5pt}
\centering
\begin{tikzpicture}[thick, scale=0.75, every node/.style={transform shape}, node distance=2cm,>=latex']
    \node [small_box_layer] (emb1) {Emb.};
    \node [left=1em of emb1] (input_1) {$\vec y_t$};
    \node [below left=1em and 1em of emb1] (input_5) {$\vec t_j$};
    \node [right=4.5em of input_5] (ctr_5) {};
    \draw [->, thick] (input_1) -- node {} (emb1);
    \node [small_box_layer, below right=0em and 1em of emb1.east] (rec) {LSTM};
    \draw [->, thick] (emb1.east) -- node {} (rec);
    \draw [->, thick, rounded corners] (input_5) -- (ctr_5.west) -- node {} (rec);
    \node [right=1em of rec] (output) {$\vec h_j$};
    \draw [->, thick] (rec) -- node {} (output);
    \node [loss_layer, right= 1em of output.east] (loss2) {$-\log f^*\left(t_{j+1}\right)$};
    \draw [->, thick] (output) -- node {} (loss2);
    \node [right=1em of loss2] (true_t) {$t_{j+1}$};
    \draw [->, thick] (true_t) -- node {} (loss2);
\end{tikzpicture}
\vspace*{-5pt}
\caption{Architecture of the recurrent neural network used in the RMTPP model to learn a representation $\vec h_j$ of an event history consisting of pairs of timings and markers $\left(t_i, y_i\right)$. $\vec t_j$ and $\vec y_j$ represent the timings and events of the history up to the $j^{\textrm{th}}$ event. $\vec h_j$ is learned through minimising the negative log-likelihood of the $j+1^{\textrm{th}}$ event occurring at time $t_{j+1}$. The hidden representation $\vec h_j$ is then used as parameter to a point process.}
\vspace*{-10pt}
\label{fig:rmtpp_diagram}
\end{figure}
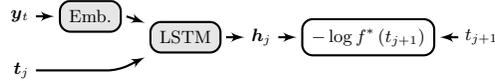

The architecture of the RNN is illustrated in Figure \ref{fig:rmtpp_diagram}. The event markers are embedded into latent space. The embedded event vector and the event timings are then fed into a recurrent layer. The recurrent layer maintains a hidden state $\vec h_j$ which summarises the event history. The recurrent layer uses a rectifier as activation function and is implemented using LSTM or GRUs.

The parameters of the recurrent layer ($\vec v^{\left(t\right)}$ and $b^{\left(t\right)}$) are learned through training the RNN using a fully connected output layer with a single neuron and linear activation, with the negative log-likelihood\footnote{The original log-likelihood function in \cite{Du2016} took into account both the likelihood of the timing of the next event and the likelihood of the marker taking certain value. The later is omitted for simplicity as we do not deal with marker prediction here.} of observing a collection of example sequences $\mathcal{C} = \left\{\left(t_j^{\left(i\right)}, y_j^{\left(i\right)}\right)_{j=1}^{n^{\left(i\right)}}\right\}_{i \in \mathbb{Z}^+}$ defined as:
\begin{align}
-\ell\left(\mathcal{C}\right) = -\sum_i \sum_j  \log f\left(t_{j+1}^{\left(i\right)}\middle|\vec h_j\right) .
\end{align}

\section{Method}
\label{sec:method}

Survival models can only accept a vector of features aggregated for each user as input. 
By using aggregated features, we discard a significant proportion of information contained in the time series of events. 
Unlike survival models, RNNs are capable of utilising the raw user history and automatically learning features. However, censored data can not be included. Omitting censored users causes predictions to be heavily biased towards low return times. 
We remove the limitations of RNNs and Cox proportional hazard models by developing a novel model that can incorporate censored data, use multiple heterogeneous markers and automatically learn features from raw time series data.

\begin{figure*}
\centering
\begin{subfigure}[t]{.32\textwidth}
  \includegraphics[width=\textwidth, trim=3.5mm 0 4.5mm 0, clip]{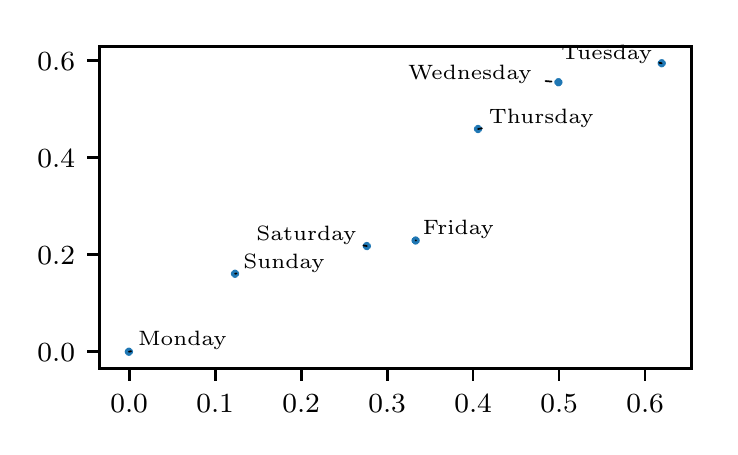}
  \vspace*{-20pt}
  \caption{Day of week embeddings (two-dimensional).}
  \label{fig:rnn_embeddings_dayOfWeek.}
\end{subfigure}
\hspace*{.003\textwidth}
\begin{subfigure}[t]{.32\textwidth}
  \includegraphics[width=\textwidth, trim=3.5mm 0 4.5mm 0, clip]{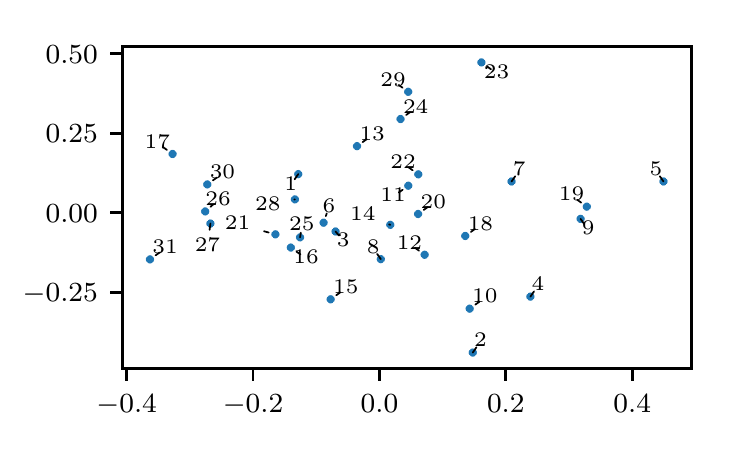}
  \vspace*{-20pt}
  \caption{Day of month embeddings (five-dimensional).}
  \label{fig:rnn_embeddings_dayOfMonth}
\end{subfigure}
\hspace*{.003\textwidth}
\begin{subfigure}[t]{.32\textwidth}
  \includegraphics[width=\textwidth, trim=3.5mm 0 4.5mm 0, clip]{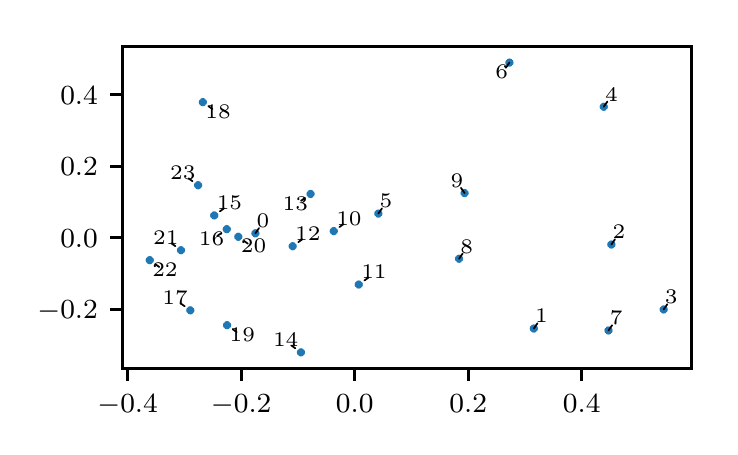}
  \vspace*{-20pt}
  \caption{Hour of day embeddings (four-dimensional).}
  \label{fig:rnn_embeddings_hourOfDay}
\end{subfigure}
\vspace*{-5pt}
\caption{Embeddings for discrete features used in the simple recurrent neural network model. The embeddings translate discrete inputs into vectors of specified dimension. We can observe a clusters of late night and early morning hours in (c), and a separation of weekend days from weekdays in (a), suggesting that viable representations are found. The embeddings are found by training a model with embeddings layers as input (alongside the remaining inputs) and then evaluating the embeddings for each possible input. Dimensionality reduction through PCA is used to produce the visualisations in case the embedding vector has more than two dimensions.
}
\vspace*{-25pt}
\label{fig:rnn_embeddings}
\end{figure*}

\subsection{Heterogeneous Markers}
In many practical settings, multiple heterogeneous markers are available describing the nature of events. Markers can be be both discrete and continuous. To encode discrete markers we use an embedding layer. We also embed cyclic features, an example being the hour of an event. Instead of encoding the hour as $\{0,1..23\}$ we learn an embedding that is able to capture the similarity between e.g. the hours 23:00 and 0:00 (see Figure~\ref{fig:rnn_embeddings} for a visualisation of the embeddings on some of the features used). Embedding layers solve this problem through mapping discrete values into a continuous vector space where similar categories are located nearby to each other. We train the network to find mappings that represent the meaning of the discrete features with respect to the task, i.e. to predict the return time.
We apply a unit norm constraint to each embedding layer, enforcing the values to be of a similar magnitude to the non-categorical variables, which are also normalized.

To avoid an expensive search for a suitable number of embedding dimensions during training, we perform a preliminary simulation. We train a model with a high number of dimensions per feature and use Principal Component Analysis (PCA) to reduce the dimensionality to the minimum number required to account for more than 90\% of the initial variance. 

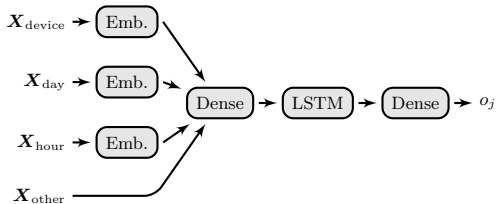
\begin{figure}
\vspace*{-10pt}
\centering
\begin{tikzpicture}[thick, scale=0.75, every node/.style={transform shape}, node distance=2cm,>=latex']
    \node [small_box_layer] (emb1) {Emb.};
    \node [small_box_layer, below=1em of emb1] (emb2) {Emb.};
    \node [small_box_layer, below=1em of emb2] (emb3) {Emb.};
    \node [left=1em of emb1] (input_1) {$\mat X_{\text{device}}$};
    \node [left=1em of emb2] (input_2) {$\mat X_{\text{day}}$};
    \node [left=1em of emb3] (input_3) {$\mat X_{\text{hour}}$};
    \node [below left=1em and 1em of emb3] (input_4) {$\mat X_{\text{other}}$};
    \node [right=4.5em of input_4] (ctr_4) {};
    \draw [->, thick] (input_1) -- node {} (emb1);
    \draw [->, thick] (input_2) -- node {} (emb2);
    \draw [->, thick] (input_3) -- node {} (emb3);
    \node [small_box_layer, below right=0em and 1em of emb2.east] (dense1) {Dense};
    \draw [->, thick] (emb1.east) -- node {} (dense1);
    \draw [->, thick] (emb2.east) -- node {} (dense1);
    \draw [->, thick] (emb3.east) -- node {} (dense1);
    \draw [->, thick, rounded corners] (input_4) -- (ctr_4.west) -- node {} (dense1);
    \node [small_box_layer, right=1em of dense1] (rec) {LSTM};
    \draw [->, thick] (dense1) -- node {} (rec);
    \node [small_box_layer, right=1em of rec] (dense) {Dense};
    \node [right=1em of dense] (output) {$o_j$};
    \draw [->, thick] (rec) -- node {} (dense);
    \draw [->, thick] (dense) -- node {} (output);
\end{tikzpicture}
\vspace*{-5pt}
\caption{Adapted recurrent marked temporal point process architecture. Embedding layers are used for each discrete feature. The embeddings and the remaining continuous features are fed into a dense layer and then the LSTM layer. The LSTM layer finds a hidden representation of the user's session history up to the $j^{\textrm{th}}$ session. A value $o_j$ is obtained from the hidden representation through a single-neuron output layer with linear activation. The negative log-likelihood of the next session occurring at time $t_{j+1}$ is used to train the model.}
\vspace*{-25pt}
\label{fig:rmtpp_adapted_diagram}
\end{figure}

The embeddings and the non-categorical features are fed into a single dense layer, which produces the input to the LSTM. Figure~\ref{fig:rmtpp_adapted_diagram} shows how the model processes heterogeneous input data.

\subsection{Recurrent Neural Network Survival Model (RNNSM)}

We combine ideas from RNNs and survival analysis to produce a model that is able to incorporate non-returning users and automatically learn features in a survival regression setting.
To achieve this we use the survival function as a factor for right-censored (non-returned user) observations. We start with the log likelihood function defined in Equation~\eqref{eq:survival_likelihood} 
\begin{align}
\begin{split}
\ell \left(\mathcal{C}\right)
  & = \sum_m \sum_{i \in \text{unc.}}\hspace{-.5em} \log f^*\left(t_{i+1}^{(m)}\right) \,
      + \sum_n\sum_{j \in \text{r.c.}}\hspace{-.3em} \log S^*\left(t_{j+1}^{( n)}\right),
\end{split}
\end{align}
which is a sum over all session intervals for all users.

The first term is thus the log-likelihood for a single returning user at time $t_{j+1}$ given an embedded representation of the user's session history up to their latest session at time $t_j$ as $\vec h_j$. This log-likelihood is defined as
\begin{align}\label{align:likelihood_return}
\begin{split}
&\ell_{\text{ret}}\left(t_{j+1}\right) = \log f^*\left(t_{j+1}\right)\\
&= o_j + w\left(t_{j+1} - t_j\right) + \frac{1}{w}\exp\left(o_j\right) - \frac{1}{w}\exp\left(o_j + w\left(t_{j+1} - t_j\right)\right),
\end{split}
\end{align}
where $o_j$ represents the output of the fully-connected layer after $j$th step of the input sequence:
$o_j = \vec v^T \vec h_j + b$.
For the expression in the second term we substitute the survival function given by
\begin{align}
S^*\left(t\right) &= \exp \left(-\int_{t_j}^t \lambda^*\left(\tau\right) d\tau\right) 
= \exp \left(\frac{1}{w}\exp \left(o_j\right) - \frac{1}{w} \exp\left(o_j + w\left(t - t_j\right)\right)\right)
\end{align}
from Equation~\eqref{align:rmtpp_time_likelihood}
to get the log-likelihood term for a single censored user:
\begin{align}
\begin{split}
\ell_{\text{non-ret}}\left(t_{j+1}\right) &= \log S^*\left(t_{j+1}\right)
= \frac{1}{w}\exp \left(o_j\right) - \frac{1}{w} \exp\left(o_j + w\left(t - t_j\right)\right) ,
\end{split}
\end{align}
where $t_{j+1}$ refers to the time between the $j$th user's last session in the observation window and the end of the prediction window.

We can now express a loss function that incorporates examples of non-returned users. Note that in a sequence of active days of a non-returning user, only the last return time is censored. 
The loss for all users is given by 
\begin{align}\label{align:adap_rmtpp_loss_seq}
-\ell(\mathcal{C}) &= -\sum_i\sum_j \ell\left(t^{(i)}_{j+1}\right) ,
\end{align}
where
\begin{align}\label{align:adap_rmtpp_lk_with_churned}
\ell\left(t_j^{(i)}\right) &= \begin{cases}
\ell_{\text{non-ret}}\left(t_j^{(i)}\right), &\text{if}\quad i\in\mathcal{C}_{\text{non-ret}}\, \text{and}\, j=n^{(i)} + 1\\
\ell_{\text{ret}}\left(t_j^{(i)}\right), & \text{otherwise}
\end{cases}
\end{align}
and $\mathcal{C} = \mathcal{C}_{\text{ret}} \,\cup\, \mathcal{C}_{\text{non-ret}}$ denotes the collections of all users' session histories, consisting of the histories of returning and non-returning users.

\subsection{Return Time Predictions}
\label{sec:cox_ph_recency}
We predict the return time, which is the time between two sessions: $d_{j+1} = t_{j+1} - t_{j}$ using the expectation of the return time given the session history:
\begin{align}\label{align:pred_t}
\hat d_{j+1} = \mathbb{E}\left[T\middle|\mathcal{H}_j\right] = \int_{t_j}^\infty S^*\left(t\right) dt .
\end{align}
This integral does not in general have a closed form solution, but can easily be evaluated using numerical integration.

However, this expression allows the model to predict users to return before the start of the prediction window (see Figure~\ref{fig:censorship}). Therefore we need to censor the predictions by finding $\mathds{E}\left[T\middle| T>t_p\right]$ where $t_p$ is the start of the prediction window. We show that the conditional expected return time can be derived from the expected return time through applying the definition of the survival function.
\begin{align}
\begin{split}
& \mathds{E}\left[T\right] = P\left(T>t_p\right) \mathds{E}\left[T\middle| T>t_p\right] + P\left(\overline{T>t_p}\right) \mathds{E}\left[T\middle| \overline{T>t_p}\right]\\
\Leftrightarrow\quad & \mathds{E}\left[T\middle| T>t_p\right] = \frac{\mathds{E}\left[T\right] - P\left(T\leq t_p\right) \mathds{E}\left[T\middle| T\leq t_p\right]}{P\left(T>t_p\right)} .
\end{split}
\end{align}

Using Equation~\eqref{eq:survival_function}, we obtain:
\begin{align}\label{align:absence_expectation}
\begin{split}
\mathds{E}\left[T\middle| T>t_s\right] &= \frac{\int_0^{\infty} S(z)\, dz - \left(1 - S\left(t_s\right)\right) \int_0^{t_s} S\left(z\right)\, dz}{S\left(t_s\right)}\\
&= \frac{\int_{t_s}^{\infty} S(z)\, dz}{S\left(t_s\right)} + \int_0^{t_s} S\left(z\right)\, dz .
\end{split}
\end{align}



\section{Experiments}
\label{sec:experiments}
Here we compare several methods for predicting user return time, discussing the advantages, assumptions and limitations of each and providing empirical results on a real-world dataset. We experiment with six distinct models: (1) a baseline model, using the time between a user's last session in the observation time frame and the beginning of the prediction time frame (``Baseline''); (2) a simple RNN architecture (``RNN'');\footnote{It consists of a single LSTM layer followed by a fully connected output layer with a single neuron. The loss function is the mean squared error using only returning users as there is no target value for non-returning users.} (3) a Cox proportional hazard model (``CPH'')  (4) a Cox proportional hazard model conditioned on absence time (``CPHA'' --- see Section~\ref{sec:cox_ph_recency}); (5) a RNN survival model (``RNNSM''); (6) a RNN survival model conditioned on absence time (``RNNSMA'' --- see Section~\ref{sec:cox_ph_recency}).

The dataset is a sample of user sessions from ASOS.com's website and mobile applications covering a period of one and a half years. Each session is associated with a user, temporal markers such as the time and duration, and behavioural data such as the number of images and videos viewed during a session. The dataset is split into training and test sets using split that is stratified to contain equal ratios of censored users. In total, there are 38,716 users in the training set and 9,680 users in the test set. 63.6\% of users in both sets return in the prediction window. In the test set, the targeted return time of returning users is 58.04 days on average with a standard deviation of 50.3 days.

\label{sec:bg_performance_metrics}
Evaluating the models based solely on the RMSE of the return time predictions is problematic because churned users can not be included. We therefore use multiple measures to compare the performance of return time models. These are the root mean squared error~\cite{Kapoor:2014:HBA:2623330.2623348,Du2016},\footnote{Unlike in the cited publications, our dataset contains a large proportion of non-returning users. The score thus only reflects the performance on returning users.} concordance index~\cite{Harrell1996MultivariableErrors},\footnote{This incorporates the knowledge of the censored return time of non-returning users.} non-returning AUC, and non-returning recall.\footnote{For AUC and recall, we treat the users who did not return within the prediction window as the positive class in a binary-classification framework.}







\begin{table*}
\vspace*{-5pt}
\centering
\begin{tabular}{lccccccccc}
\toprule
& \multicolumn{1}{c}{Baseline} & \multicolumn{1}{c}{{CPH}} & \multicolumn{1}{c}{CPHA} & \multicolumn{1}{c}{{RNN}} & \multicolumn{1}{c}{{RNNSM}} & \multicolumn{1}{c}{RNNSMA} \\
\midrule
RMSE (days) & 43.25 & 49.99 & 59.81 & \textbf{28.69} & 59.99 & 63.76\\
Concordance & 0.500 & 0.816 & \textbf{0.817} & 0.706 & 0.739 & 0.740\\
Non-returning AUC & 0.743 & 0.793 & 0.788 & 0.763 & \textbf{0.796} & 0.794\\
Non-returning recall  & 0.000 & 0.246 & 0.461 & 0.000 & 0.538 & \textbf{0.605}\\
\bottomrule
\end{tabular}
\vspace{5pt}
\caption{Comparison of performance of return time prediction models. The RMSE is just for returning users. Best values for each performance metric are highlighted in bold.}
\vspace*{-20pt}
\label{tab:performance_metrics_comp}
\end{table*}

\subsection{Result on Performance Metrics}

We report the performance metrics on the test set after training the models in Table~\ref{tab:performance_metrics_comp}. As the test dataset only contains users that were active in the activity window, the baseline model predicts that all users will return.

The CPH model uses an aggregated representation of each user's session history and additional metrics such as the absence time. The CPH model outperforms the baseline in each performance metric with the exception of the RMSE. This suggests that the improved non-returning recall rate can therefore be partially attributed to the model learning a positive bias for return times. This effect is even more pronounced for the CPHA model. However, the improvement in the concordance score demonstrates that, beyond a positive bias, a better relative ordering of predictions is achieved. Both CPH models perform particularly well in terms of the concordance score, suggesting that their predictions best reflect the relative ordering.

The RNN model cannot recall any non-returning users as its training examples only included returning users. However, the RMSE score demonstrates that the RNN model is superior in terms of predicting the return time of returning users and thus that sequential information is predictive of return time. 

Finally, the recurrent neural network-based survival model (RNNSM) further improves the recall of non-returning users over the CPHA model without notable changes in the RMSE. More importantly it obtains the best performance for non-returning AUC, meaning it is the best model to discriminate between returning and non-returning users in the prediction window. Applying the absence-conditioned expectation to obtain predictions from the RNNSM further improves the model's performance on non-returning recall. However, the concordance scores of both RNNSM models suggest that the relative ordering is not reflected as well as by the CPH model.

\subsection{Prediction error in relation to true return time}
To evaluate the performance of each model in more detail we group users by their true return time, rounded down to a week. We then evaluate each model's performance based on a number of error metrics. This is to assess the usefulness of each model; for example, a model which performs well on short return times but poorly on long return times would be less useful in practice than one that performs equally well on a wide distribution of return times.

\begin{figure*}
\vspace*{-10pt}
\centering
\begin{subfigure}{.325\textwidth}
  \centering
  \hspace*{-12pt}
  \includegraphics[width=\textwidth, trim=2mm 0 3mm 0, clip]{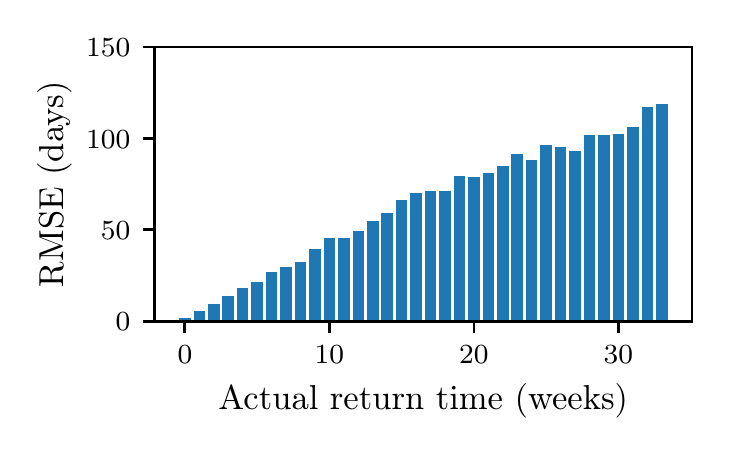}
  \vspace*{-10pt}
  \caption{Baseline}
  \label{fig:rmsebywk_baseline_absence}
\end{subfigure}
\begin{subfigure}{.325\textwidth}
  \centering
  \hspace*{-8pt}
  \includegraphics[width=\textwidth, trim=2mm 0 3mm 0, clip]{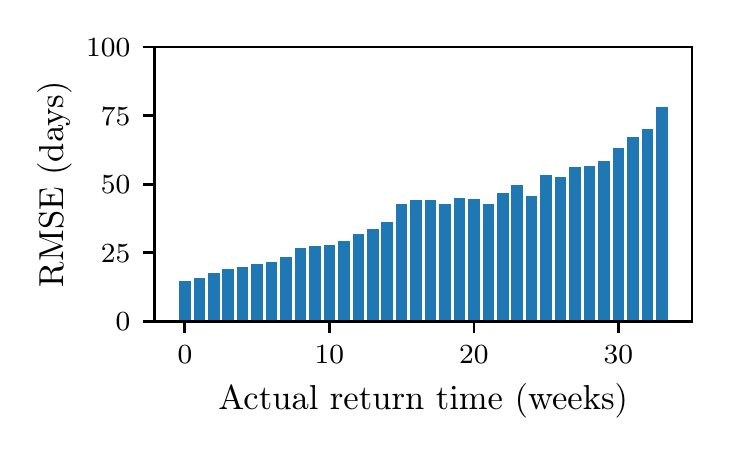}
  \vspace*{-10pt}
  \caption{Simple RNN}
  \label{fig:rmsebywk_rnn}
\end{subfigure}
\begin{subfigure}{.325\textwidth}
  \centering
  \hspace*{-6pt}
  \includegraphics[width=\textwidth, trim=2mm 0 3mm 0, clip]{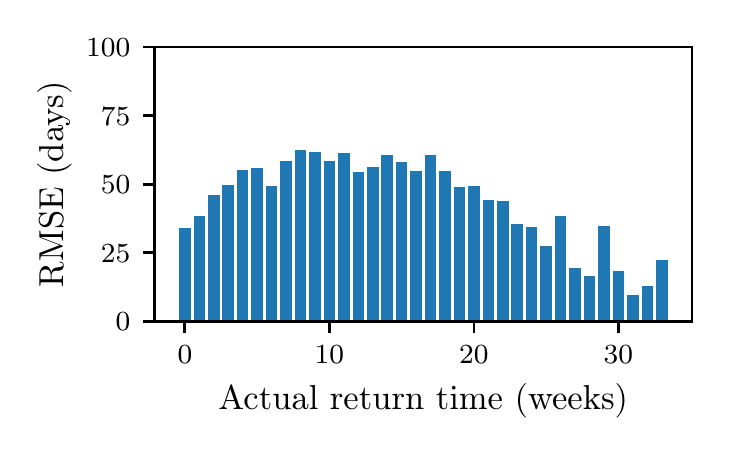}
  \vspace*{-10pt}
  \caption{CPH}
  \label{fig:rmsebywk_cox_ph_noabs}
\end{subfigure}
\begin{subfigure}{.325\textwidth}
  \centering
  \hspace*{-12pt}
  \includegraphics[width=\textwidth, trim=2mm 0 3mm 0, clip]{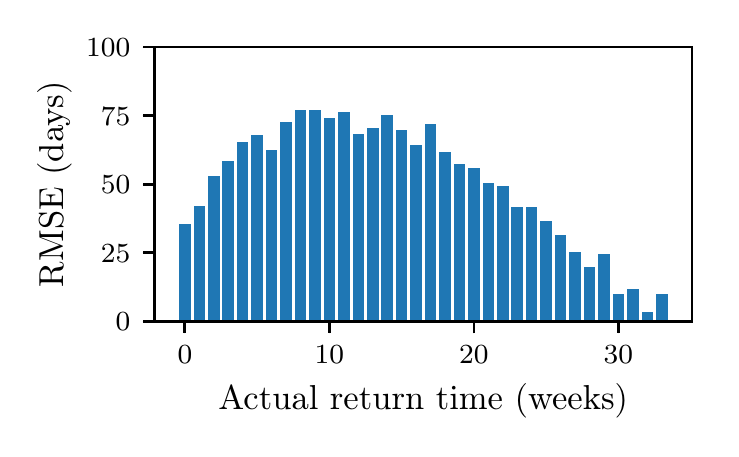}
  \vspace*{-10pt}
  \caption{CPHA}
  \label{fig:rmsebywk_cox_ph_abs}
\end{subfigure}
\begin{subfigure}{.325\textwidth}
  \centering
  \hspace*{-8pt}
  \includegraphics[width=\textwidth, trim=2mm 0 3mm 0, clip]{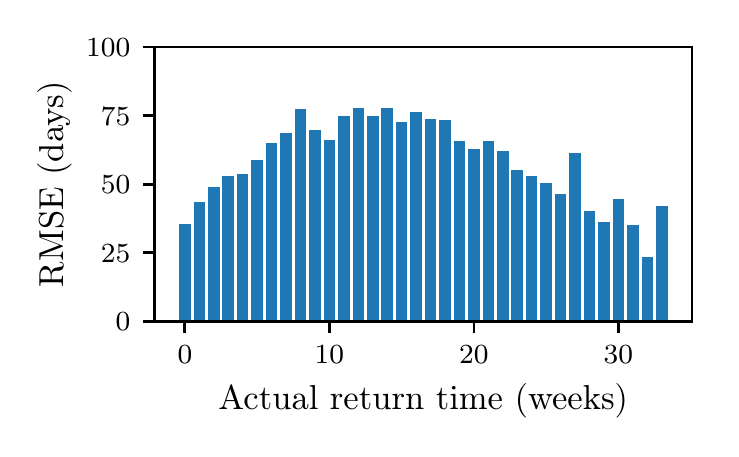}
  \vspace*{-10pt}
  \caption{RNNSM}
  \label{fig:rmsebywk_rmtpp_noabs}
\end{subfigure}
\begin{subfigure}{.325\textwidth}
  \centering
  \hspace*{-6pt}
  \includegraphics[width=\textwidth, trim=2mm 0 3mm 0, clip]{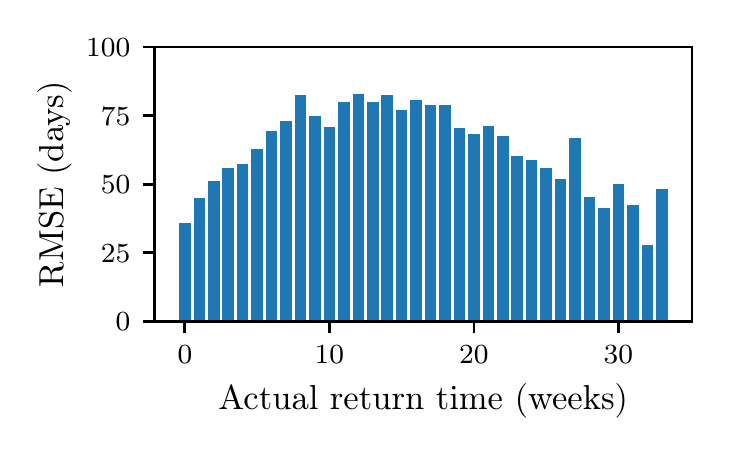}
  \vspace*{-10pt}
  \caption{RNNSMA}
  \label{fig:rmsebywk_rmtpp_abs}
\end{subfigure}
\vspace*{-5pt}
\caption{Root mean squared error (RMSE) in relation to true return time compared between return time prediction models. In order to get a more detailed impression of each model's performance in relation to the true return time, we group users by their true return time rounded down to weeks. For each group, we then find the RMSE. Note the adjusted scale for the baseline model (a).}
\vspace*{-12pt}
\label{fig:rmsebywk}
\end{figure*}

\paragraph{Root mean squared error}
The RMSE in relation to the return time in weeks is shown for each return time prediction model in Figure~\ref{fig:rmsebywk}. 
The majority of users in the dataset return within ten weeks; we see that for the baseline model and the RNN model the RMSE for these users is relatively low, this gives a low overall RMSE. However, for users who have longer return times both of those models perform increasingly poorly for increasing true return times.

For the models that incorporate training examples of both non-returning and returning users we see a different pattern. The performance for users with longer return times is generally better than those returning earlier. This demonstrates that these models are able to use censored observations to improve predictions for returning users. While the overall RMSE is lower for the CPH model compared the RNNSM (see Table~\ref{tab:performance_metrics_comp}), the distribution of errors is skewed, with a higher RMSE for earlier returning users.

We also see the effect of the absence-conditioned return time expectation. For the CPH model there is an increase in performance for users with very long return times, however there is a significant negative impact on users with shorter return times. This results suggest that the absence-conditioned expectation is more suitable for the RNNSM as it seems to have little effect on the RMSE distribution whilst improving the non-returning recall as can be seen in Table~\ref{tab:performance_metrics_comp}.

\begin{figure*}
\centering
\begin{subfigure}{.325\textwidth}
  \centering
  \includegraphics[width=\textwidth, trim=2mm 0 3mm 0, clip]{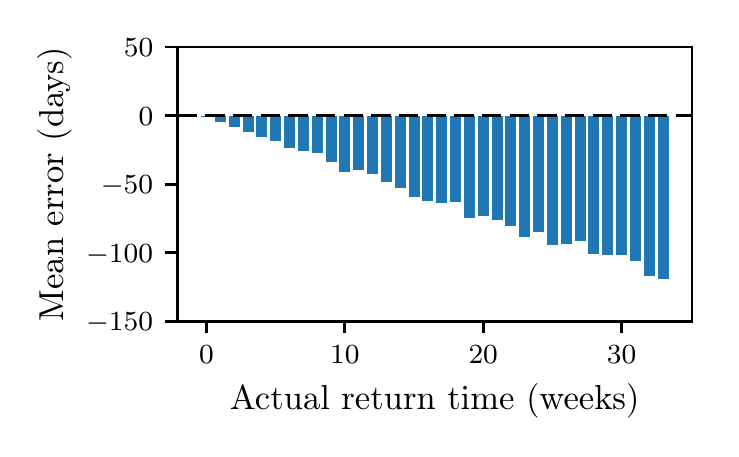}
  \vspace*{-20pt}
  \caption{Baseline}
  \label{fig:baseline_absence_errbywk}
\end{subfigure}
\begin{subfigure}{.325\textwidth}
  \centering
  \includegraphics[width=\textwidth, trim=2mm 0 3mm 0, clip]{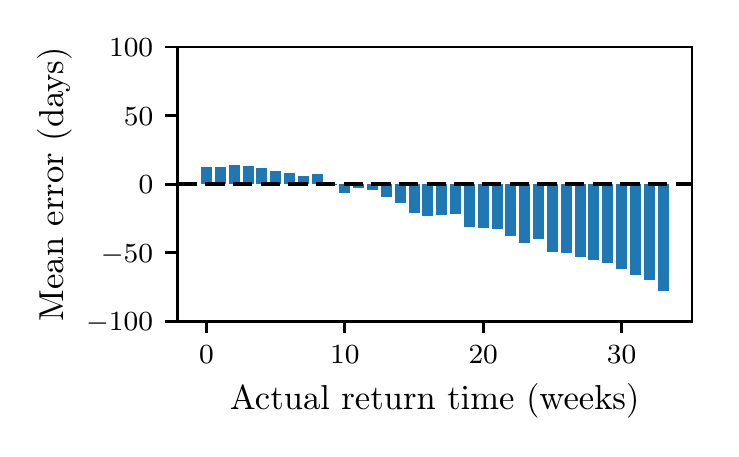}
  \vspace*{-20pt}
  \caption{RNN}
  \label{fig:rnn_errbywk}
\end{subfigure}
\begin{subfigure}{.325\textwidth}
  \centering
  \includegraphics[width=\textwidth, trim=2mm 0 3mm 0, clip]{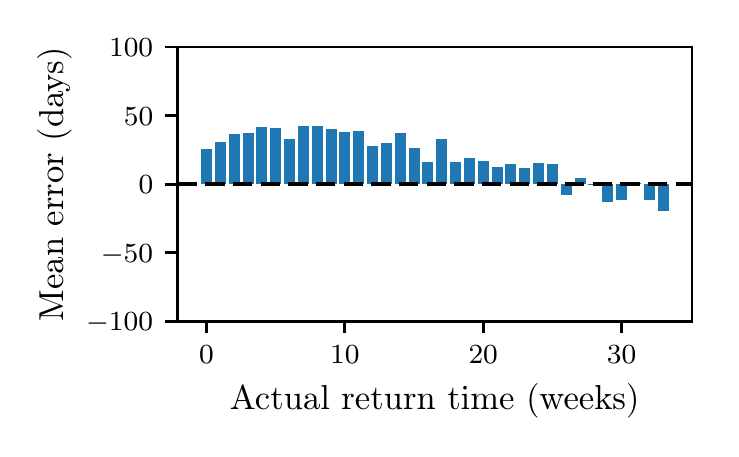}
  \vspace*{-20pt}
  \caption{CPH}
  \label{fig:cox_ph_noabs_errbywk}
\end{subfigure}\\
\begin{subfigure}{.325\textwidth}
  \centering
  \includegraphics[width=\textwidth, trim=2mm 0 3mm 0, clip]{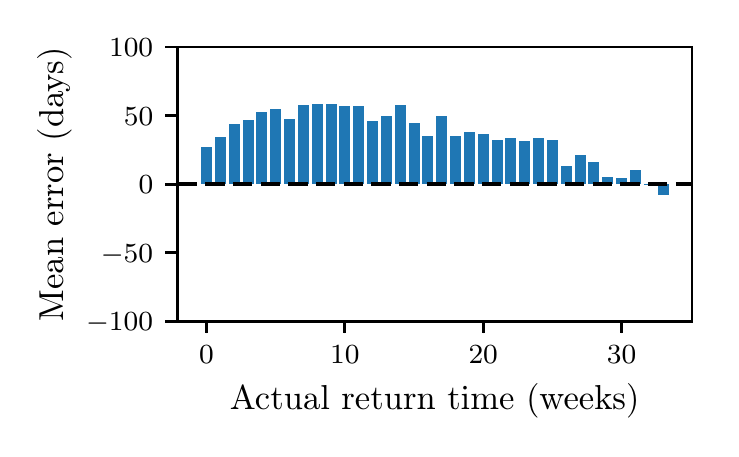}
  \vspace*{-20pt}
  \caption{CPHA}
  \label{fig:cox_ph_abs_errbywk}
\end{subfigure}
\begin{subfigure}{.325\textwidth}
  \centering
  \includegraphics[width=\textwidth, trim=2mm 0 3mm 0, clip]{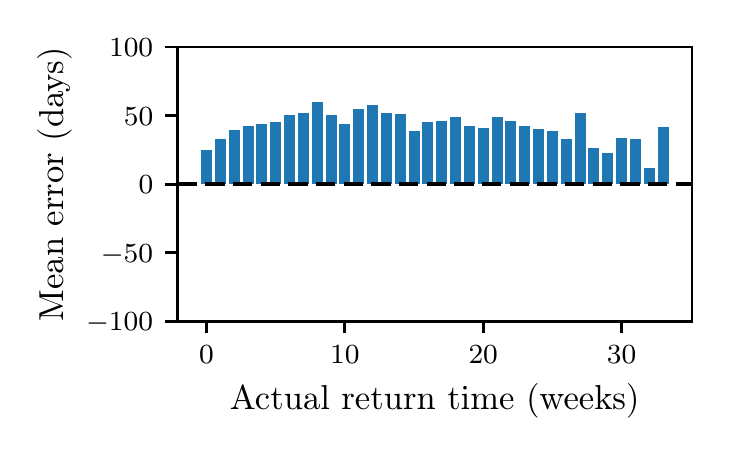}
  \vspace*{-20pt}
  \caption{RNNSM}
  \label{fig:rmtpp_noabs_errbywk}
\end{subfigure}
\begin{subfigure}{.325\textwidth}
  \centering
  \includegraphics[width=\textwidth, trim=2mm 0 3mm 0, clip]{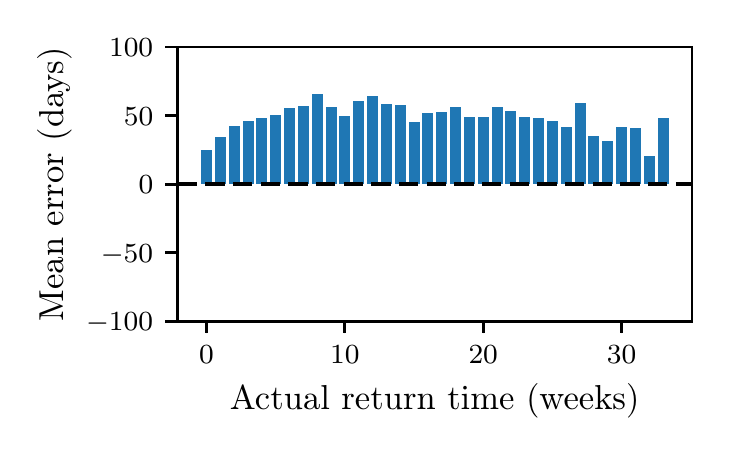}
  \vspace*{-20pt}
  \caption{RNNSMA}
  \label{fig:rmtpp_abs_errbywk}
\end{subfigure}
\vspace*{-5pt}
\caption{Mean error per in relation to true return time comparison for all return time prediction models. Users are grouped by their true return time to the nearest week. This allows us to determine how the prediction bias is related to the true return time. We see that models which include non-returning users in training have a positive bias on the prediction for returning users, those that don't have a negative bias. Note the adjusted range for the baseline model (a).}
\vspace*{-10pt}
\label{fig:errbywk}
\end{figure*}

\paragraph{Mean error}
Figure~\ref{fig:errbywk} shows the mean error for each group and each model. The baseline model will always underestimate the true return time as by definition it can only predict a value equal to or lower than the true return time. The RNN model's
performance is worse for users returning later, this is due to the restriction of predicting all users to return within a certain window leading to a negative bias for users returning later. The CPH model and the RNNSMs both overestimate the return times of the majority of users. It is possible to subtract the mean prediction error on the training set from these predictions in order to reduce the error, however this would lead to a reduction in non-returning AUC as overall return times would be reduced. 

\subsection{Error in relation to number of active days}\label{sec:err_vs_days}
\begin{figure}
\vspace*{-10pt}
\centering
\begin{subfigure}{.45\textwidth}
  \centering
  \includegraphics[width=\textwidth]{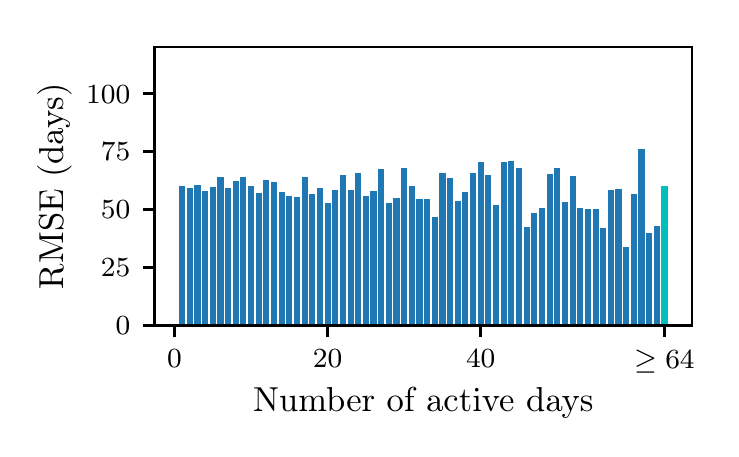}
  \vspace*{-20pt}
  \caption{CPHA}
  \label{fig:rmsenumsess_cox_ph_abs}
\end{subfigure}
\begin{subfigure}{.45\textwidth}
  \centering
  \includegraphics[width=\textwidth]{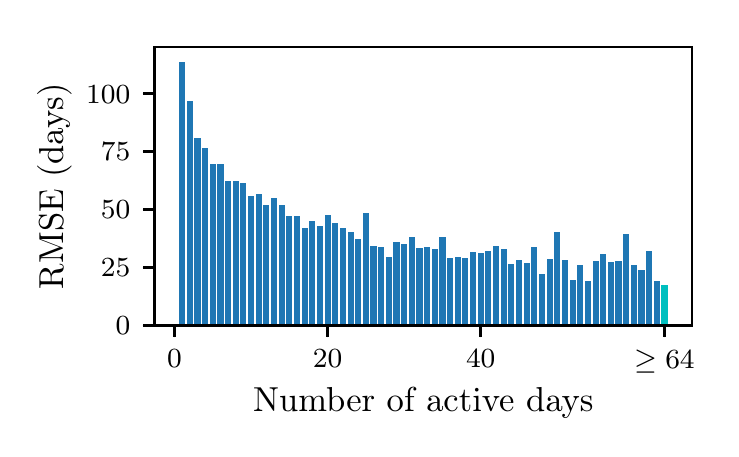}
  \vspace*{-20pt}
  \caption{RNNSMA}
  \label{fig:rmsenumsess_rmtpp_abs}
\end{subfigure}
  \vspace*{-8pt}
\caption{The number of days a user is active for does not affect the prediction quality of the CPHA model, while a greater number of active days improves the performance of the RNNSMA. The last bar in both charts represents all users with 64 or more active days.}
\vspace*{-10pt}
\label{fig:rmsenumsess}
\end{figure}

In this section we group users by their number of active days, an active day is a day on which a user had a session. We plot the RMSE in days for the CPHA model and RNNSMA model in Figure~\ref{fig:rmsenumsess}. These are the two best performing models which include returning users in terms of non-returning AUC and recall. We use up to 64 active days per user -- we therefore group all users with 64 or more active days. We can immediately see that the RNNSM is able to make better predictions for users with a a higher number of active days. This is not the case for the CPHA model. This demonstrates that for users with more active days (longer RNN input sequences) the RNNSM model improves greatly. This again indicates that the sequence information captured by the RNNSM is predictive of user return and is preferable for users with a larger number of sessions. 

\section{Discussion}

We have developed the RNNSM, a novel model that overcomes the weaknesses of survival models and RNNs for user return time prediction. We have highlighted the importance of including right-censored observations in return time prediction models and extended the Cox proportional hazard model to include users' absence time. We found that for modelling recurring events, a limitation of existing survival regression models is that they only operate on aggregate representations instead of raw time-series data. We addressed this problem by using a RNN point process model, which combines the advantages of survival regression and overcomes the limitation of RNNs by including censored observations. We extended the RMTPP model to include any number of session markers and developed a method of training the model using censored observations. We further demonstrated how to include users' absence times. The RNNSM successfully learns from sequences of sessions and outperforms all other models in predicting which users are not going to return (non-returning AUC).



\bibliographystyle{splncs04}
\providecommand{\url}[1]{\texttt{#1}}
\providecommand{\urlprefix}{URL }

\bibliography{return_time_pred_short}

\end{document}

%% file: censorship.pgf
\begingroup%
\makeatletter%
\begin{pgfpicture}%
\pgfpathrectangle{\pgfpointorigin}{\pgfqpoint{5.900000in}{2.950000in}}%
\pgfusepath{use as bounding box, clip}%
\begin{pgfscope}%
\pgfsetbuttcap%
\pgfsetmiterjoin%
\definecolor{currentfill}{rgb}{1.000000,1.000000,1.000000}%
\pgfsetfillcolor{currentfill}%
\pgfsetlinewidth{0.000000pt}%
\definecolor{currentstroke}{rgb}{1.000000,1.000000,1.000000}%
\pgfsetstrokecolor{currentstroke}%
\pgfsetdash{}{0pt}%
\pgfpathmoveto{\pgfqpoint{0.000000in}{0.000000in}}%
\pgfpathlineto{\pgfqpoint{5.900000in}{0.000000in}}%
\pgfpathlineto{\pgfqpoint{5.900000in}{2.950000in}}%
\pgfpathlineto{\pgfqpoint{0.000000in}{2.950000in}}%
\pgfpathclose%
\pgfusepath{fill}%
\end{pgfscope}%
\begin{pgfscope}%
\pgfsetbuttcap%
\pgfsetmiterjoin%
\definecolor{currentfill}{rgb}{1.000000,1.000000,1.000000}%
\pgfsetfillcolor{currentfill}%
\pgfsetlinewidth{0.000000pt}%
\definecolor{currentstroke}{rgb}{0.000000,0.000000,0.000000}%
\pgfsetstrokecolor{currentstroke}%
\pgfsetstrokeopacity{0.000000}%
\pgfsetdash{}{0pt}%
\pgfpathmoveto{\pgfqpoint{1.010460in}{0.354000in}}%
\pgfpathlineto{\pgfqpoint{5.546000in}{0.354000in}}%
\pgfpathlineto{\pgfqpoint{5.546000in}{2.773000in}}%
\pgfpathlineto{\pgfqpoint{1.010460in}{2.773000in}}%
\pgfpathclose%
\pgfusepath{fill}%
\end{pgfscope}%
\begin{pgfscope}%
\pgfpathrectangle{\pgfqpoint{1.010460in}{0.354000in}}{\pgfqpoint{4.535540in}{2.419000in}} %
\pgfusepath{clip}%
\pgfsetbuttcap%
\pgfsetroundjoin%
\definecolor{currentfill}{rgb}{0.839216,0.152941,0.156863}%
\pgfsetfillcolor{currentfill}%
\pgfsetlinewidth{1.003750pt}%
\definecolor{currentstroke}{rgb}{0.839216,0.152941,0.156863}%
\pgfsetstrokecolor{currentstroke}%
\pgfsetdash{}{0pt}%
\pgfpathmoveto{\pgfqpoint{3.517993in}{0.954802in}}%
\pgfpathlineto{\pgfqpoint{3.576919in}{1.013727in}}%
\pgfpathlineto{\pgfqpoint{3.517993in}{1.072653in}}%
\pgfpathlineto{\pgfqpoint{3.459067in}{1.013727in}}%
\pgfpathclose%
\pgfusepath{stroke,fill}%
\end{pgfscope}%
\begin{pgfscope}%
\pgfpathrectangle{\pgfqpoint{1.010460in}{0.354000in}}{\pgfqpoint{4.535540in}{2.419000in}} %
\pgfusepath{clip}%
\pgfsetbuttcap%
\pgfsetroundjoin%
\definecolor{currentfill}{rgb}{0.839216,0.152941,0.156863}%
\pgfsetfillcolor{currentfill}%
\pgfsetlinewidth{1.003750pt}%
\definecolor{currentstroke}{rgb}{0.839216,0.152941,0.156863}%
\pgfsetstrokecolor{currentstroke}%
\pgfsetdash{}{0pt}%
\pgfpathmoveto{\pgfqpoint{2.479563in}{1.614529in}}%
\pgfpathlineto{\pgfqpoint{2.538489in}{1.673455in}}%
\pgfpathlineto{\pgfqpoint{2.479563in}{1.732380in}}%
\pgfpathlineto{\pgfqpoint{2.420638in}{1.673455in}}%
\pgfpathclose%
\pgfusepath{stroke,fill}%
\end{pgfscope}%
\begin{pgfscope}%
\pgfpathrectangle{\pgfqpoint{1.010460in}{0.354000in}}{\pgfqpoint{4.535540in}{2.419000in}} %
\pgfusepath{clip}%
\pgfsetbuttcap%
\pgfsetroundjoin%
\definecolor{currentfill}{rgb}{0.839216,0.152941,0.156863}%
\pgfsetfillcolor{currentfill}%
\pgfsetlinewidth{1.003750pt}%
\definecolor{currentstroke}{rgb}{0.839216,0.152941,0.156863}%
\pgfsetstrokecolor{currentstroke}%
\pgfsetdash{}{0pt}%
\pgfpathmoveto{\pgfqpoint{3.073651in}{2.494165in}}%
\pgfpathlineto{\pgfqpoint{3.132577in}{2.553091in}}%
\pgfpathlineto{\pgfqpoint{3.073651in}{2.612016in}}%
\pgfpathlineto{\pgfqpoint{3.014726in}{2.553091in}}%
\pgfpathclose%
\pgfusepath{stroke,fill}%
\end{pgfscope}%
\begin{pgfscope}%
\pgfpathrectangle{\pgfqpoint{1.010460in}{0.354000in}}{\pgfqpoint{4.535540in}{2.419000in}} %
\pgfusepath{clip}%
\pgfsetbuttcap%
\pgfsetmiterjoin%
\definecolor{currentfill}{rgb}{0.172549,0.627451,0.172549}%
\pgfsetfillcolor{currentfill}%
\pgfsetfillopacity{0.200000}%
\pgfsetlinewidth{0.000000pt}%
\definecolor{currentstroke}{rgb}{0.000000,0.000000,0.000000}%
\pgfsetstrokecolor{currentstroke}%
\pgfsetstrokeopacity{0.200000}%
\pgfsetdash{}{0pt}%
\pgfpathmoveto{\pgfqpoint{3.543079in}{0.354000in}}%
\pgfpathlineto{\pgfqpoint{3.543079in}{2.773000in}}%
\pgfpathlineto{\pgfqpoint{5.546000in}{2.773000in}}%
\pgfpathlineto{\pgfqpoint{5.546000in}{0.354000in}}%
\pgfpathclose%
\pgfusepath{fill}%
\end{pgfscope}%
\begin{pgfscope}%
\pgfsetbuttcap%
\pgfsetroundjoin%
\definecolor{currentfill}{rgb}{0.000000,0.000000,0.000000}%
\pgfsetfillcolor{currentfill}%
\pgfsetlinewidth{0.803000pt}%
\definecolor{currentstroke}{rgb}{0.000000,0.000000,0.000000}%
\pgfsetstrokecolor{currentstroke}%
\pgfsetdash{}{0pt}%
\pgfsys@defobject{currentmarker}{\pgfqpoint{0.000000in}{-0.048611in}}{\pgfqpoint{0.000000in}{0.000000in}}{%
\pgfpathmoveto{\pgfqpoint{0.000000in}{0.000000in}}%
\pgfpathlineto{\pgfqpoint{0.000000in}{-0.048611in}}%
\pgfusepath{stroke,fill}%
}%
\begin{pgfscope}%
\pgfsys@transformshift{1.507052in}{0.354000in}%
\pgfsys@useobject{currentmarker}{}%
\end{pgfscope}%
\end{pgfscope}%
\begin{pgfscope}%
\end{pgfscope}%
\begin{pgfscope}%
\pgfsetbuttcap%
\pgfsetroundjoin%
\definecolor{currentfill}{rgb}{0.000000,0.000000,0.000000}%
\pgfsetfillcolor{currentfill}%
\pgfsetlinewidth{0.803000pt}%
\definecolor{currentstroke}{rgb}{0.000000,0.000000,0.000000}%
\pgfsetstrokecolor{currentstroke}%
\pgfsetdash{}{0pt}%
\pgfsys@defobject{currentmarker}{\pgfqpoint{0.000000in}{-0.048611in}}{\pgfqpoint{0.000000in}{0.000000in}}{%
\pgfpathmoveto{\pgfqpoint{0.000000in}{0.000000in}}%
\pgfpathlineto{\pgfqpoint{0.000000in}{-0.048611in}}%
\pgfusepath{stroke,fill}%
}%
\begin{pgfscope}%
\pgfsys@transformshift{2.516789in}{0.354000in}%
\pgfsys@useobject{currentmarker}{}%
\end{pgfscope}%
\end{pgfscope}%
\begin{pgfscope}%
\end{pgfscope}%
\begin{pgfscope}%
\pgfsetbuttcap%
\pgfsetroundjoin%
\definecolor{currentfill}{rgb}{0.000000,0.000000,0.000000}%
\pgfsetfillcolor{currentfill}%
\pgfsetlinewidth{0.803000pt}%
\definecolor{currentstroke}{rgb}{0.000000,0.000000,0.000000}%
\pgfsetstrokecolor{currentstroke}%
\pgfsetdash{}{0pt}%
\pgfsys@defobject{currentmarker}{\pgfqpoint{0.000000in}{-0.048611in}}{\pgfqpoint{0.000000in}{0.000000in}}{%
\pgfpathmoveto{\pgfqpoint{0.000000in}{0.000000in}}%
\pgfpathlineto{\pgfqpoint{0.000000in}{-0.048611in}}%
\pgfusepath{stroke,fill}%
}%
\begin{pgfscope}%
\pgfsys@transformshift{3.543079in}{0.354000in}%
\pgfsys@useobject{currentmarker}{}%
\end{pgfscope}%
\end{pgfscope}%
\begin{pgfscope}%
\end{pgfscope}%
\begin{pgfscope}%
\pgfsetbuttcap%
\pgfsetroundjoin%
\definecolor{currentfill}{rgb}{0.000000,0.000000,0.000000}%
\pgfsetfillcolor{currentfill}%
\pgfsetlinewidth{0.803000pt}%
\definecolor{currentstroke}{rgb}{0.000000,0.000000,0.000000}%
\pgfsetstrokecolor{currentstroke}%
\pgfsetdash{}{0pt}%
\pgfsys@defobject{currentmarker}{\pgfqpoint{0.000000in}{-0.048611in}}{\pgfqpoint{0.000000in}{0.000000in}}{%
\pgfpathmoveto{\pgfqpoint{0.000000in}{0.000000in}}%
\pgfpathlineto{\pgfqpoint{0.000000in}{-0.048611in}}%
\pgfusepath{stroke,fill}%
}%
\begin{pgfscope}%
\pgfsys@transformshift{4.536263in}{0.354000in}%
\pgfsys@useobject{currentmarker}{}%
\end{pgfscope}%
\end{pgfscope}%
\begin{pgfscope}%
\end{pgfscope}%
\begin{pgfscope}%
\pgfsetbuttcap%
\pgfsetroundjoin%
\definecolor{currentfill}{rgb}{0.000000,0.000000,0.000000}%
\pgfsetfillcolor{currentfill}%
\pgfsetlinewidth{0.803000pt}%
\definecolor{currentstroke}{rgb}{0.000000,0.000000,0.000000}%
\pgfsetstrokecolor{currentstroke}%
\pgfsetdash{}{0pt}%
\pgfsys@defobject{currentmarker}{\pgfqpoint{0.000000in}{-0.048611in}}{\pgfqpoint{0.000000in}{0.000000in}}{%
\pgfpathmoveto{\pgfqpoint{0.000000in}{0.000000in}}%
\pgfpathlineto{\pgfqpoint{0.000000in}{-0.048611in}}%
\pgfusepath{stroke,fill}%
}%
\begin{pgfscope}%
\pgfsys@transformshift{5.546000in}{0.354000in}%
\pgfsys@useobject{currentmarker}{}%
\end{pgfscope}%
\end{pgfscope}%
\begin{pgfscope}%
\end{pgfscope}%
\begin{pgfscope}%
\pgfsetrectcap%
\pgfsetroundjoin%
\pgfsetlinewidth{1.505625pt}%
\definecolor{currentstroke}{rgb}{0.000000,0.000000,0.000000}%
\pgfsetstrokecolor{currentstroke}%
\pgfsetdash{}{0pt}%
\pgfpathmoveto{\pgfqpoint{1.001508in}{0.317715in}}%
\pgfpathlineto{\pgfqpoint{1.019411in}{0.390285in}}%
\pgfusepath{stroke}%
\end{pgfscope}%
\begin{pgfscope}%
\pgfsetrectcap%
\pgfsetroundjoin%
\pgfsetlinewidth{1.505625pt}%
\definecolor{currentstroke}{rgb}{0.000000,0.000000,0.000000}%
\pgfsetstrokecolor{currentstroke}%
\pgfsetdash{}{0pt}%
\pgfpathmoveto{\pgfqpoint{1.001508in}{2.736715in}}%
\pgfpathlineto{\pgfqpoint{1.019411in}{2.809285in}}%
\pgfusepath{stroke}%
\end{pgfscope}%
\begin{pgfscope}%
\pgfpathrectangle{\pgfqpoint{1.010460in}{0.354000in}}{\pgfqpoint{4.535540in}{2.419000in}} %
\pgfusepath{clip}%
\pgfsetrectcap%
\pgfsetroundjoin%
\pgfsetlinewidth{1.505625pt}%
\definecolor{currentstroke}{rgb}{0.121569,0.466667,0.705882}%
\pgfsetstrokecolor{currentstroke}%
\pgfsetdash{}{0pt}%
\pgfpathmoveto{\pgfqpoint{3.337463in}{0.573909in}}%
\pgfpathlineto{\pgfqpoint{4.316557in}{0.573909in}}%
\pgfusepath{stroke}%
\end{pgfscope}%
\begin{pgfscope}%
\pgfpathrectangle{\pgfqpoint{1.010460in}{0.354000in}}{\pgfqpoint{4.535540in}{2.419000in}} %
\pgfusepath{clip}%
\pgfsetbuttcap%
\pgfsetmiterjoin%
\definecolor{currentfill}{rgb}{0.121569,0.466667,0.705882}%
\pgfsetfillcolor{currentfill}%
\pgfsetlinewidth{1.003750pt}%
\definecolor{currentstroke}{rgb}{0.121569,0.466667,0.705882}%
\pgfsetstrokecolor{currentstroke}%
\pgfsetdash{}{0pt}%
\pgfsys@defobject{currentmarker}{\pgfqpoint{-0.058926in}{-0.058926in}}{\pgfqpoint{0.058926in}{0.058926in}}{%
\pgfpathmoveto{\pgfqpoint{-0.000000in}{-0.058926in}}%
\pgfpathlineto{\pgfqpoint{0.058926in}{0.000000in}}%
\pgfpathlineto{\pgfqpoint{0.000000in}{0.058926in}}%
\pgfpathlineto{\pgfqpoint{-0.058926in}{0.000000in}}%
\pgfpathclose%
\pgfusepath{stroke,fill}%
}%
\begin{pgfscope}%
\pgfsys@transformshift{3.337463in}{0.573909in}%
\pgfsys@useobject{currentmarker}{}%
\end{pgfscope}%
\begin{pgfscope}%
\pgfsys@transformshift{4.316557in}{0.573909in}%
\pgfsys@useobject{currentmarker}{}%
\end{pgfscope}%
\end{pgfscope}%
\begin{pgfscope}%
\pgfpathrectangle{\pgfqpoint{1.010460in}{0.354000in}}{\pgfqpoint{4.535540in}{2.419000in}} %
\pgfusepath{clip}%
\pgfsetbuttcap%
\pgfsetroundjoin%
\pgfsetlinewidth{1.505625pt}%
\definecolor{currentstroke}{rgb}{0.121569,0.466667,0.705882}%
\pgfsetstrokecolor{currentstroke}%
\pgfsetstrokeopacity{0.100000}%
\pgfsetdash{{2.200000pt}{2.200000pt}}{0.000000pt}%
\pgfpathmoveto{\pgfqpoint{1.005460in}{0.573909in}}%
\pgfpathlineto{\pgfqpoint{3.337463in}{0.573909in}}%
\pgfusepath{stroke}%
\end{pgfscope}%
\begin{pgfscope}%
\pgfpathrectangle{\pgfqpoint{1.010460in}{0.354000in}}{\pgfqpoint{4.535540in}{2.419000in}} %
\pgfusepath{clip}%
\pgfsetrectcap%
\pgfsetroundjoin%
\pgfsetlinewidth{1.505625pt}%
\definecolor{currentstroke}{rgb}{0.121569,0.466667,0.705882}%
\pgfsetstrokecolor{currentstroke}%
\pgfsetdash{}{0pt}%
\pgfpathmoveto{\pgfqpoint{3.270290in}{0.793818in}}%
\pgfpathlineto{\pgfqpoint{3.589885in}{0.793818in}}%
\pgfusepath{stroke}%
\end{pgfscope}%
\begin{pgfscope}%
\pgfpathrectangle{\pgfqpoint{1.010460in}{0.354000in}}{\pgfqpoint{4.535540in}{2.419000in}} %
\pgfusepath{clip}%
\pgfsetbuttcap%
\pgfsetmiterjoin%
\definecolor{currentfill}{rgb}{0.121569,0.466667,0.705882}%
\pgfsetfillcolor{currentfill}%
\pgfsetlinewidth{1.003750pt}%
\definecolor{currentstroke}{rgb}{0.121569,0.466667,0.705882}%
\pgfsetstrokecolor{currentstroke}%
\pgfsetdash{}{0pt}%
\pgfsys@defobject{currentmarker}{\pgfqpoint{-0.058926in}{-0.058926in}}{\pgfqpoint{0.058926in}{0.058926in}}{%
\pgfpathmoveto{\pgfqpoint{-0.000000in}{-0.058926in}}%
\pgfpathlineto{\pgfqpoint{0.058926in}{0.000000in}}%
\pgfpathlineto{\pgfqpoint{0.000000in}{0.058926in}}%
\pgfpathlineto{\pgfqpoint{-0.058926in}{0.000000in}}%
\pgfpathclose%
\pgfusepath{stroke,fill}%
}%
\begin{pgfscope}%
\pgfsys@transformshift{3.270290in}{0.793818in}%
\pgfsys@useobject{currentmarker}{}%
\end{pgfscope}%
\begin{pgfscope}%
\pgfsys@transformshift{3.589885in}{0.793818in}%
\pgfsys@useobject{currentmarker}{}%
\end{pgfscope}%
\end{pgfscope}%
\begin{pgfscope}%
\pgfpathrectangle{\pgfqpoint{1.010460in}{0.354000in}}{\pgfqpoint{4.535540in}{2.419000in}} %
\pgfusepath{clip}%
\pgfsetbuttcap%
\pgfsetroundjoin%
\pgfsetlinewidth{1.505625pt}%
\definecolor{currentstroke}{rgb}{0.121569,0.466667,0.705882}%
\pgfsetstrokecolor{currentstroke}%
\pgfsetstrokeopacity{0.100000}%
\pgfsetdash{{2.200000pt}{2.200000pt}}{0.000000pt}%
\pgfpathmoveto{\pgfqpoint{1.005460in}{0.793818in}}%
\pgfpathlineto{\pgfqpoint{3.270290in}{0.793818in}}%
\pgfusepath{stroke}%
\end{pgfscope}%
\begin{pgfscope}%
\pgfpathrectangle{\pgfqpoint{1.010460in}{0.354000in}}{\pgfqpoint{4.535540in}{2.419000in}} %
\pgfusepath{clip}%
\pgfsetbuttcap%
\pgfsetroundjoin%
\pgfsetlinewidth{1.505625pt}%
\definecolor{currentstroke}{rgb}{0.839216,0.152941,0.156863}%
\pgfsetstrokecolor{currentstroke}%
\pgfsetstrokeopacity{0.100000}%
\pgfsetdash{{2.200000pt}{2.200000pt}}{0.000000pt}%
\pgfpathmoveto{\pgfqpoint{1.005460in}{1.013727in}}%
\pgfpathlineto{\pgfqpoint{3.517993in}{1.013727in}}%
\pgfusepath{stroke}%
\end{pgfscope}%
\begin{pgfscope}%
\pgfpathrectangle{\pgfqpoint{1.010460in}{0.354000in}}{\pgfqpoint{4.535540in}{2.419000in}} %
\pgfusepath{clip}%
\pgfsetrectcap%
\pgfsetroundjoin%
\pgfsetlinewidth{1.505625pt}%
\definecolor{currentstroke}{rgb}{0.839216,0.152941,0.156863}%
\pgfsetstrokecolor{currentstroke}%
\pgfsetdash{}{0pt}%
\pgfpathmoveto{\pgfqpoint{3.517993in}{1.013727in}}%
\pgfpathlineto{\pgfqpoint{5.546000in}{1.013727in}}%
\pgfusepath{stroke}%
\end{pgfscope}%
\begin{pgfscope}%
\pgfpathrectangle{\pgfqpoint{1.010460in}{0.354000in}}{\pgfqpoint{4.535540in}{2.419000in}} %
\pgfusepath{clip}%
\pgfsetrectcap%
\pgfsetroundjoin%
\pgfsetlinewidth{1.505625pt}%
\definecolor{currentstroke}{rgb}{0.121569,0.466667,0.705882}%
\pgfsetstrokecolor{currentstroke}%
\pgfsetdash{}{0pt}%
\pgfpathmoveto{\pgfqpoint{3.390013in}{1.233636in}}%
\pgfpathlineto{\pgfqpoint{3.553312in}{1.233636in}}%
\pgfusepath{stroke}%
\end{pgfscope}%
\begin{pgfscope}%
\pgfpathrectangle{\pgfqpoint{1.010460in}{0.354000in}}{\pgfqpoint{4.535540in}{2.419000in}} %
\pgfusepath{clip}%
\pgfsetbuttcap%
\pgfsetmiterjoin%
\definecolor{currentfill}{rgb}{0.121569,0.466667,0.705882}%
\pgfsetfillcolor{currentfill}%
\pgfsetlinewidth{1.003750pt}%
\definecolor{currentstroke}{rgb}{0.121569,0.466667,0.705882}%
\pgfsetstrokecolor{currentstroke}%
\pgfsetdash{}{0pt}%
\pgfsys@defobject{currentmarker}{\pgfqpoint{-0.058926in}{-0.058926in}}{\pgfqpoint{0.058926in}{0.058926in}}{%
\pgfpathmoveto{\pgfqpoint{-0.000000in}{-0.058926in}}%
\pgfpathlineto{\pgfqpoint{0.058926in}{0.000000in}}%
\pgfpathlineto{\pgfqpoint{0.000000in}{0.058926in}}%
\pgfpathlineto{\pgfqpoint{-0.058926in}{0.000000in}}%
\pgfpathclose%
\pgfusepath{stroke,fill}%
}%
\begin{pgfscope}%
\pgfsys@transformshift{3.390013in}{1.233636in}%
\pgfsys@useobject{currentmarker}{}%
\end{pgfscope}%
\begin{pgfscope}%
\pgfsys@transformshift{3.553312in}{1.233636in}%
\pgfsys@useobject{currentmarker}{}%
\end{pgfscope}%
\end{pgfscope}%
\begin{pgfscope}%
\pgfpathrectangle{\pgfqpoint{1.010460in}{0.354000in}}{\pgfqpoint{4.535540in}{2.419000in}} %
\pgfusepath{clip}%
\pgfsetbuttcap%
\pgfsetroundjoin%
\pgfsetlinewidth{1.505625pt}%
\definecolor{currentstroke}{rgb}{0.121569,0.466667,0.705882}%
\pgfsetstrokecolor{currentstroke}%
\pgfsetstrokeopacity{0.100000}%
\pgfsetdash{{2.200000pt}{2.200000pt}}{0.000000pt}%
\pgfpathmoveto{\pgfqpoint{1.005460in}{1.233636in}}%
\pgfpathlineto{\pgfqpoint{3.390013in}{1.233636in}}%
\pgfusepath{stroke}%
\end{pgfscope}%
\begin{pgfscope}%
\pgfpathrectangle{\pgfqpoint{1.010460in}{0.354000in}}{\pgfqpoint{4.535540in}{2.419000in}} %
\pgfusepath{clip}%
\pgfsetrectcap%
\pgfsetroundjoin%
\pgfsetlinewidth{1.505625pt}%
\definecolor{currentstroke}{rgb}{0.121569,0.466667,0.705882}%
\pgfsetstrokecolor{currentstroke}%
\pgfsetdash{}{0pt}%
\pgfpathmoveto{\pgfqpoint{2.379230in}{1.453545in}}%
\pgfpathlineto{\pgfqpoint{5.390318in}{1.453545in}}%
\pgfusepath{stroke}%
\end{pgfscope}%
\begin{pgfscope}%
\pgfpathrectangle{\pgfqpoint{1.010460in}{0.354000in}}{\pgfqpoint{4.535540in}{2.419000in}} %
\pgfusepath{clip}%
\pgfsetbuttcap%
\pgfsetmiterjoin%
\definecolor{currentfill}{rgb}{0.121569,0.466667,0.705882}%
\pgfsetfillcolor{currentfill}%
\pgfsetlinewidth{1.003750pt}%
\definecolor{currentstroke}{rgb}{0.121569,0.466667,0.705882}%
\pgfsetstrokecolor{currentstroke}%
\pgfsetdash{}{0pt}%
\pgfsys@defobject{currentmarker}{\pgfqpoint{-0.058926in}{-0.058926in}}{\pgfqpoint{0.058926in}{0.058926in}}{%
\pgfpathmoveto{\pgfqpoint{-0.000000in}{-0.058926in}}%
\pgfpathlineto{\pgfqpoint{0.058926in}{0.000000in}}%
\pgfpathlineto{\pgfqpoint{0.000000in}{0.058926in}}%
\pgfpathlineto{\pgfqpoint{-0.058926in}{0.000000in}}%
\pgfpathclose%
\pgfusepath{stroke,fill}%
}%
\begin{pgfscope}%
\pgfsys@transformshift{2.379230in}{1.453545in}%
\pgfsys@useobject{currentmarker}{}%
\end{pgfscope}%
\begin{pgfscope}%
\pgfsys@transformshift{5.390318in}{1.453545in}%
\pgfsys@useobject{currentmarker}{}%
\end{pgfscope}%
\end{pgfscope}%
\begin{pgfscope}%
\pgfpathrectangle{\pgfqpoint{1.010460in}{0.354000in}}{\pgfqpoint{4.535540in}{2.419000in}} %
\pgfusepath{clip}%
\pgfsetbuttcap%
\pgfsetroundjoin%
\pgfsetlinewidth{1.505625pt}%
\definecolor{currentstroke}{rgb}{0.121569,0.466667,0.705882}%
\pgfsetstrokecolor{currentstroke}%
\pgfsetstrokeopacity{0.100000}%
\pgfsetdash{{2.200000pt}{2.200000pt}}{0.000000pt}%
\pgfpathmoveto{\pgfqpoint{1.005460in}{1.453545in}}%
\pgfpathlineto{\pgfqpoint{2.379230in}{1.453545in}}%
\pgfusepath{stroke}%
\end{pgfscope}%
\begin{pgfscope}%
\pgfpathrectangle{\pgfqpoint{1.010460in}{0.354000in}}{\pgfqpoint{4.535540in}{2.419000in}} %
\pgfusepath{clip}%
\pgfsetbuttcap%
\pgfsetroundjoin%
\pgfsetlinewidth{1.505625pt}%
\definecolor{currentstroke}{rgb}{0.839216,0.152941,0.156863}%
\pgfsetstrokecolor{currentstroke}%
\pgfsetstrokeopacity{0.100000}%
\pgfsetdash{{2.200000pt}{2.200000pt}}{0.000000pt}%
\pgfpathmoveto{\pgfqpoint{1.005460in}{1.673455in}}%
\pgfpathlineto{\pgfqpoint{2.479563in}{1.673455in}}%
\pgfusepath{stroke}%
\end{pgfscope}%
\begin{pgfscope}%
\pgfpathrectangle{\pgfqpoint{1.010460in}{0.354000in}}{\pgfqpoint{4.535540in}{2.419000in}} %
\pgfusepath{clip}%
\pgfsetrectcap%
\pgfsetroundjoin%
\pgfsetlinewidth{1.505625pt}%
\definecolor{currentstroke}{rgb}{0.839216,0.152941,0.156863}%
\pgfsetstrokecolor{currentstroke}%
\pgfsetdash{}{0pt}%
\pgfpathmoveto{\pgfqpoint{2.479563in}{1.673455in}}%
\pgfpathlineto{\pgfqpoint{5.546000in}{1.673455in}}%
\pgfusepath{stroke}%
\end{pgfscope}%
\begin{pgfscope}%
\pgfpathrectangle{\pgfqpoint{1.010460in}{0.354000in}}{\pgfqpoint{4.535540in}{2.419000in}} %
\pgfusepath{clip}%
\pgfsetrectcap%
\pgfsetroundjoin%
\pgfsetlinewidth{1.505625pt}%
\definecolor{currentstroke}{rgb}{0.121569,0.466667,0.705882}%
\pgfsetstrokecolor{currentstroke}%
\pgfsetdash{}{0pt}%
\pgfpathmoveto{\pgfqpoint{2.995529in}{1.893364in}}%
\pgfpathlineto{\pgfqpoint{4.824704in}{1.893364in}}%
\pgfusepath{stroke}%
\end{pgfscope}%
\begin{pgfscope}%
\pgfpathrectangle{\pgfqpoint{1.010460in}{0.354000in}}{\pgfqpoint{4.535540in}{2.419000in}} %
\pgfusepath{clip}%
\pgfsetbuttcap%
\pgfsetmiterjoin%
\definecolor{currentfill}{rgb}{0.121569,0.466667,0.705882}%
\pgfsetfillcolor{currentfill}%
\pgfsetlinewidth{1.003750pt}%
\definecolor{currentstroke}{rgb}{0.121569,0.466667,0.705882}%
\pgfsetstrokecolor{currentstroke}%
\pgfsetdash{}{0pt}%
\pgfsys@defobject{currentmarker}{\pgfqpoint{-0.058926in}{-0.058926in}}{\pgfqpoint{0.058926in}{0.058926in}}{%
\pgfpathmoveto{\pgfqpoint{-0.000000in}{-0.058926in}}%
\pgfpathlineto{\pgfqpoint{0.058926in}{0.000000in}}%
\pgfpathlineto{\pgfqpoint{0.000000in}{0.058926in}}%
\pgfpathlineto{\pgfqpoint{-0.058926in}{0.000000in}}%
\pgfpathclose%
\pgfusepath{stroke,fill}%
}%
\begin{pgfscope}%
\pgfsys@transformshift{2.995529in}{1.893364in}%
\pgfsys@useobject{currentmarker}{}%
\end{pgfscope}%
\begin{pgfscope}%
\pgfsys@transformshift{4.824704in}{1.893364in}%
\pgfsys@useobject{currentmarker}{}%
\end{pgfscope}%
\end{pgfscope}%
\begin{pgfscope}%
\pgfpathrectangle{\pgfqpoint{1.010460in}{0.354000in}}{\pgfqpoint{4.535540in}{2.419000in}} %
\pgfusepath{clip}%
\pgfsetbuttcap%
\pgfsetroundjoin%
\pgfsetlinewidth{1.505625pt}%
\definecolor{currentstroke}{rgb}{0.121569,0.466667,0.705882}%
\pgfsetstrokecolor{currentstroke}%
\pgfsetstrokeopacity{0.100000}%
\pgfsetdash{{2.200000pt}{2.200000pt}}{0.000000pt}%
\pgfpathmoveto{\pgfqpoint{1.005460in}{1.893364in}}%
\pgfpathlineto{\pgfqpoint{2.995529in}{1.893364in}}%
\pgfusepath{stroke}%
\end{pgfscope}%
\begin{pgfscope}%
\pgfpathrectangle{\pgfqpoint{1.010460in}{0.354000in}}{\pgfqpoint{4.535540in}{2.419000in}} %
\pgfusepath{clip}%
\pgfsetrectcap%
\pgfsetroundjoin%
\pgfsetlinewidth{1.505625pt}%
\definecolor{currentstroke}{rgb}{0.121569,0.466667,0.705882}%
\pgfsetstrokecolor{currentstroke}%
\pgfsetdash{}{0pt}%
\pgfpathmoveto{\pgfqpoint{3.491633in}{2.113273in}}%
\pgfpathlineto{\pgfqpoint{3.953642in}{2.113273in}}%
\pgfusepath{stroke}%
\end{pgfscope}%
\begin{pgfscope}%
\pgfpathrectangle{\pgfqpoint{1.010460in}{0.354000in}}{\pgfqpoint{4.535540in}{2.419000in}} %
\pgfusepath{clip}%
\pgfsetbuttcap%
\pgfsetmiterjoin%
\definecolor{currentfill}{rgb}{0.121569,0.466667,0.705882}%
\pgfsetfillcolor{currentfill}%
\pgfsetlinewidth{1.003750pt}%
\definecolor{currentstroke}{rgb}{0.121569,0.466667,0.705882}%
\pgfsetstrokecolor{currentstroke}%
\pgfsetdash{}{0pt}%
\pgfsys@defobject{currentmarker}{\pgfqpoint{-0.058926in}{-0.058926in}}{\pgfqpoint{0.058926in}{0.058926in}}{%
\pgfpathmoveto{\pgfqpoint{-0.000000in}{-0.058926in}}%
\pgfpathlineto{\pgfqpoint{0.058926in}{0.000000in}}%
\pgfpathlineto{\pgfqpoint{0.000000in}{0.058926in}}%
\pgfpathlineto{\pgfqpoint{-0.058926in}{0.000000in}}%
\pgfpathclose%
\pgfusepath{stroke,fill}%
}%
\begin{pgfscope}%
\pgfsys@transformshift{3.491633in}{2.113273in}%
\pgfsys@useobject{currentmarker}{}%
\end{pgfscope}%
\begin{pgfscope}%
\pgfsys@transformshift{3.953642in}{2.113273in}%
\pgfsys@useobject{currentmarker}{}%
\end{pgfscope}%
\end{pgfscope}%
\begin{pgfscope}%
\pgfpathrectangle{\pgfqpoint{1.010460in}{0.354000in}}{\pgfqpoint{4.535540in}{2.419000in}} %
\pgfusepath{clip}%
\pgfsetbuttcap%
\pgfsetroundjoin%
\pgfsetlinewidth{1.505625pt}%
\definecolor{currentstroke}{rgb}{0.121569,0.466667,0.705882}%
\pgfsetstrokecolor{currentstroke}%
\pgfsetstrokeopacity{0.100000}%
\pgfsetdash{{2.200000pt}{2.200000pt}}{0.000000pt}%
\pgfpathmoveto{\pgfqpoint{1.005460in}{2.113273in}}%
\pgfpathlineto{\pgfqpoint{3.491633in}{2.113273in}}%
\pgfusepath{stroke}%
\end{pgfscope}%
\begin{pgfscope}%
\pgfpathrectangle{\pgfqpoint{1.010460in}{0.354000in}}{\pgfqpoint{4.535540in}{2.419000in}} %
\pgfusepath{clip}%
\pgfsetrectcap%
\pgfsetroundjoin%
\pgfsetlinewidth{1.505625pt}%
\definecolor{currentstroke}{rgb}{0.121569,0.466667,0.705882}%
\pgfsetstrokecolor{currentstroke}%
\pgfsetdash{}{0pt}%
\pgfpathmoveto{\pgfqpoint{1.844663in}{2.333182in}}%
\pgfpathlineto{\pgfqpoint{4.761000in}{2.333182in}}%
\pgfusepath{stroke}%
\end{pgfscope}%
\begin{pgfscope}%
\pgfpathrectangle{\pgfqpoint{1.010460in}{0.354000in}}{\pgfqpoint{4.535540in}{2.419000in}} %
\pgfusepath{clip}%
\pgfsetbuttcap%
\pgfsetmiterjoin%
\definecolor{currentfill}{rgb}{0.121569,0.466667,0.705882}%
\pgfsetfillcolor{currentfill}%
\pgfsetlinewidth{1.003750pt}%
\definecolor{currentstroke}{rgb}{0.121569,0.466667,0.705882}%
\pgfsetstrokecolor{currentstroke}%
\pgfsetdash{}{0pt}%
\pgfsys@defobject{currentmarker}{\pgfqpoint{-0.058926in}{-0.058926in}}{\pgfqpoint{0.058926in}{0.058926in}}{%
\pgfpathmoveto{\pgfqpoint{-0.000000in}{-0.058926in}}%
\pgfpathlineto{\pgfqpoint{0.058926in}{0.000000in}}%
\pgfpathlineto{\pgfqpoint{0.000000in}{0.058926in}}%
\pgfpathlineto{\pgfqpoint{-0.058926in}{0.000000in}}%
\pgfpathclose%
\pgfusepath{stroke,fill}%
}%
\begin{pgfscope}%
\pgfsys@transformshift{1.844663in}{2.333182in}%
\pgfsys@useobject{currentmarker}{}%
\end{pgfscope}%
\begin{pgfscope}%
\pgfsys@transformshift{4.761000in}{2.333182in}%
\pgfsys@useobject{currentmarker}{}%
\end{pgfscope}%
\end{pgfscope}%
\begin{pgfscope}%
\pgfpathrectangle{\pgfqpoint{1.010460in}{0.354000in}}{\pgfqpoint{4.535540in}{2.419000in}} %
\pgfusepath{clip}%
\pgfsetbuttcap%
\pgfsetroundjoin%
\pgfsetlinewidth{1.505625pt}%
\definecolor{currentstroke}{rgb}{0.121569,0.466667,0.705882}%
\pgfsetstrokecolor{currentstroke}%
\pgfsetstrokeopacity{0.100000}%
\pgfsetdash{{2.200000pt}{2.200000pt}}{0.000000pt}%
\pgfpathmoveto{\pgfqpoint{1.005460in}{2.333182in}}%
\pgfpathlineto{\pgfqpoint{1.844663in}{2.333182in}}%
\pgfusepath{stroke}%
\end{pgfscope}%
\begin{pgfscope}%
\pgfpathrectangle{\pgfqpoint{1.010460in}{0.354000in}}{\pgfqpoint{4.535540in}{2.419000in}} %
\pgfusepath{clip}%
\pgfsetbuttcap%
\pgfsetroundjoin%
\pgfsetlinewidth{1.505625pt}%
\definecolor{currentstroke}{rgb}{0.839216,0.152941,0.156863}%
\pgfsetstrokecolor{currentstroke}%
\pgfsetstrokeopacity{0.100000}%
\pgfsetdash{{2.200000pt}{2.200000pt}}{0.000000pt}%
\pgfpathmoveto{\pgfqpoint{1.005460in}{2.553091in}}%
\pgfpathlineto{\pgfqpoint{3.073651in}{2.553091in}}%
\pgfusepath{stroke}%
\end{pgfscope}%
\begin{pgfscope}%
\pgfpathrectangle{\pgfqpoint{1.010460in}{0.354000in}}{\pgfqpoint{4.535540in}{2.419000in}} %
\pgfusepath{clip}%
\pgfsetrectcap%
\pgfsetroundjoin%
\pgfsetlinewidth{1.505625pt}%
\definecolor{currentstroke}{rgb}{0.839216,0.152941,0.156863}%
\pgfsetstrokecolor{currentstroke}%
\pgfsetdash{}{0pt}%
\pgfpathmoveto{\pgfqpoint{3.073651in}{2.553091in}}%
\pgfpathlineto{\pgfqpoint{5.546000in}{2.553091in}}%
\pgfusepath{stroke}%
\end{pgfscope}%
\begin{pgfscope}%
\pgfsetrectcap%
\pgfsetmiterjoin%
\pgfsetlinewidth{0.803000pt}%
\definecolor{currentstroke}{rgb}{0.000000,0.000000,0.000000}%
\pgfsetstrokecolor{currentstroke}%
\pgfsetdash{}{0pt}%
\pgfpathmoveto{\pgfqpoint{5.546000in}{0.354000in}}%
\pgfpathlineto{\pgfqpoint{5.546000in}{2.773000in}}%
\pgfusepath{stroke}%
\end{pgfscope}%
\begin{pgfscope}%
\pgfsetrectcap%
\pgfsetmiterjoin%
\pgfsetlinewidth{0.803000pt}%
\definecolor{currentstroke}{rgb}{0.000000,0.000000,0.000000}%
\pgfsetstrokecolor{currentstroke}%
\pgfsetdash{}{0pt}%
\pgfpathmoveto{\pgfqpoint{1.010460in}{0.354000in}}%
\pgfpathlineto{\pgfqpoint{5.546000in}{0.354000in}}%
\pgfusepath{stroke}%
\end{pgfscope}%
\begin{pgfscope}%
\pgfsetrectcap%
\pgfsetmiterjoin%
\pgfsetlinewidth{0.803000pt}%
\definecolor{currentstroke}{rgb}{0.000000,0.000000,0.000000}%
\pgfsetstrokecolor{currentstroke}%
\pgfsetdash{}{0pt}%
\pgfpathmoveto{\pgfqpoint{1.010460in}{2.773000in}}%
\pgfpathlineto{\pgfqpoint{5.546000in}{2.773000in}}%
\pgfusepath{stroke}%
\end{pgfscope}%
\begin{pgfscope}%
\pgfsetbuttcap%
\pgfsetmiterjoin%
\definecolor{currentfill}{rgb}{1.000000,1.000000,1.000000}%
\pgfsetfillcolor{currentfill}%
\pgfsetlinewidth{0.000000pt}%
\definecolor{currentstroke}{rgb}{0.000000,0.000000,0.000000}%
\pgfsetstrokecolor{currentstroke}%
\pgfsetstrokeopacity{0.000000}%
\pgfsetdash{}{0pt}%
\pgfpathmoveto{\pgfqpoint{0.354000in}{0.354000in}}%
\pgfpathlineto{\pgfqpoint{0.950782in}{0.354000in}}%
\pgfpathlineto{\pgfqpoint{0.950782in}{2.773000in}}%
\pgfpathlineto{\pgfqpoint{0.354000in}{2.773000in}}%
\pgfpathclose%
\pgfusepath{fill}%
\end{pgfscope}%
\begin{pgfscope}%
\pgfsetbuttcap%
\pgfsetroundjoin%
\definecolor{currentfill}{rgb}{0.000000,0.000000,0.000000}%
\pgfsetfillcolor{currentfill}%
\pgfsetlinewidth{0.803000pt}%
\definecolor{currentstroke}{rgb}{0.000000,0.000000,0.000000}%
\pgfsetstrokecolor{currentstroke}%
\pgfsetdash{}{0pt}%
\pgfsys@defobject{currentmarker}{\pgfqpoint{0.000000in}{-0.048611in}}{\pgfqpoint{0.000000in}{0.000000in}}{%
\pgfpathmoveto{\pgfqpoint{0.000000in}{0.000000in}}%
\pgfpathlineto{\pgfqpoint{0.000000in}{-0.048611in}}%
\pgfusepath{stroke,fill}%
}%
\begin{pgfscope}%
\pgfsys@transformshift{0.354000in}{0.354000in}%
\pgfsys@useobject{currentmarker}{}%
\end{pgfscope}%
\end{pgfscope}%
\begin{pgfscope}%
\end{pgfscope}%
\begin{pgfscope}%
\pgfsetrectcap%
\pgfsetroundjoin%
\pgfsetlinewidth{1.505625pt}%
\definecolor{currentstroke}{rgb}{0.000000,0.000000,0.000000}%
\pgfsetstrokecolor{currentstroke}%
\pgfsetdash{}{0pt}%
\pgfpathmoveto{\pgfqpoint{0.941830in}{0.317715in}}%
\pgfpathlineto{\pgfqpoint{0.959733in}{0.390285in}}%
\pgfusepath{stroke}%
\end{pgfscope}%
\begin{pgfscope}%
\pgfsetrectcap%
\pgfsetroundjoin%
\pgfsetlinewidth{1.505625pt}%
\definecolor{currentstroke}{rgb}{0.000000,0.000000,0.000000}%
\pgfsetstrokecolor{currentstroke}%
\pgfsetdash{}{0pt}%
\pgfpathmoveto{\pgfqpoint{0.941830in}{2.736715in}}%
\pgfpathlineto{\pgfqpoint{0.959733in}{2.809285in}}%
\pgfusepath{stroke}%
\end{pgfscope}%
\begin{pgfscope}%
\pgfpathrectangle{\pgfqpoint{0.354000in}{0.354000in}}{\pgfqpoint{0.596782in}{2.419000in}} %
\pgfusepath{clip}%
\pgfsetbuttcap%
\pgfsetroundjoin%
\pgfsetlinewidth{1.505625pt}%
\definecolor{currentstroke}{rgb}{0.121569,0.466667,0.705882}%
\pgfsetstrokecolor{currentstroke}%
\pgfsetstrokeopacity{0.100000}%
\pgfsetdash{{2.200000pt}{2.200000pt}}{0.000000pt}%
\pgfpathmoveto{\pgfqpoint{0.354000in}{0.573909in}}%
\pgfpathlineto{\pgfqpoint{0.955782in}{0.573909in}}%
\pgfusepath{stroke}%
\end{pgfscope}%
\begin{pgfscope}%
\pgfpathrectangle{\pgfqpoint{0.354000in}{0.354000in}}{\pgfqpoint{0.596782in}{2.419000in}} %
\pgfusepath{clip}%
\pgfsetbuttcap%
\pgfsetroundjoin%
\pgfsetlinewidth{1.505625pt}%
\definecolor{currentstroke}{rgb}{0.121569,0.466667,0.705882}%
\pgfsetstrokecolor{currentstroke}%
\pgfsetstrokeopacity{0.100000}%
\pgfsetdash{{2.200000pt}{2.200000pt}}{0.000000pt}%
\pgfpathmoveto{\pgfqpoint{0.354000in}{0.793818in}}%
\pgfpathlineto{\pgfqpoint{0.955782in}{0.793818in}}%
\pgfusepath{stroke}%
\end{pgfscope}%
\begin{pgfscope}%
\pgfpathrectangle{\pgfqpoint{0.354000in}{0.354000in}}{\pgfqpoint{0.596782in}{2.419000in}} %
\pgfusepath{clip}%
\pgfsetbuttcap%
\pgfsetroundjoin%
\pgfsetlinewidth{1.505625pt}%
\definecolor{currentstroke}{rgb}{0.839216,0.152941,0.156863}%
\pgfsetstrokecolor{currentstroke}%
\pgfsetstrokeopacity{0.100000}%
\pgfsetdash{{2.200000pt}{2.200000pt}}{0.000000pt}%
\pgfpathmoveto{\pgfqpoint{0.354000in}{1.013727in}}%
\pgfpathlineto{\pgfqpoint{0.955782in}{1.013727in}}%
\pgfusepath{stroke}%
\end{pgfscope}%
\begin{pgfscope}%
\pgfpathrectangle{\pgfqpoint{0.354000in}{0.354000in}}{\pgfqpoint{0.596782in}{2.419000in}} %
\pgfusepath{clip}%
\pgfsetbuttcap%
\pgfsetroundjoin%
\pgfsetlinewidth{1.505625pt}%
\definecolor{currentstroke}{rgb}{0.121569,0.466667,0.705882}%
\pgfsetstrokecolor{currentstroke}%
\pgfsetstrokeopacity{0.100000}%
\pgfsetdash{{2.200000pt}{2.200000pt}}{0.000000pt}%
\pgfpathmoveto{\pgfqpoint{0.354000in}{1.233636in}}%
\pgfpathlineto{\pgfqpoint{0.955782in}{1.233636in}}%
\pgfusepath{stroke}%
\end{pgfscope}%
\begin{pgfscope}%
\pgfpathrectangle{\pgfqpoint{0.354000in}{0.354000in}}{\pgfqpoint{0.596782in}{2.419000in}} %
\pgfusepath{clip}%
\pgfsetbuttcap%
\pgfsetroundjoin%
\pgfsetlinewidth{1.505625pt}%
\definecolor{currentstroke}{rgb}{0.121569,0.466667,0.705882}%
\pgfsetstrokecolor{currentstroke}%
\pgfsetstrokeopacity{0.100000}%
\pgfsetdash{{2.200000pt}{2.200000pt}}{0.000000pt}%
\pgfpathmoveto{\pgfqpoint{0.354000in}{1.453545in}}%
\pgfpathlineto{\pgfqpoint{0.955782in}{1.453545in}}%
\pgfusepath{stroke}%
\end{pgfscope}%
\begin{pgfscope}%
\pgfpathrectangle{\pgfqpoint{0.354000in}{0.354000in}}{\pgfqpoint{0.596782in}{2.419000in}} %
\pgfusepath{clip}%
\pgfsetbuttcap%
\pgfsetroundjoin%
\pgfsetlinewidth{1.505625pt}%
\definecolor{currentstroke}{rgb}{0.839216,0.152941,0.156863}%
\pgfsetstrokecolor{currentstroke}%
\pgfsetstrokeopacity{0.100000}%
\pgfsetdash{{2.200000pt}{2.200000pt}}{0.000000pt}%
\pgfpathmoveto{\pgfqpoint{0.354000in}{1.673455in}}%
\pgfpathlineto{\pgfqpoint{0.955782in}{1.673455in}}%
\pgfusepath{stroke}%
\end{pgfscope}%
\begin{pgfscope}%
\pgfpathrectangle{\pgfqpoint{0.354000in}{0.354000in}}{\pgfqpoint{0.596782in}{2.419000in}} %
\pgfusepath{clip}%
\pgfsetbuttcap%
\pgfsetroundjoin%
\pgfsetlinewidth{1.505625pt}%
\definecolor{currentstroke}{rgb}{0.121569,0.466667,0.705882}%
\pgfsetstrokecolor{currentstroke}%
\pgfsetstrokeopacity{0.100000}%
\pgfsetdash{{2.200000pt}{2.200000pt}}{0.000000pt}%
\pgfpathmoveto{\pgfqpoint{0.354000in}{1.893364in}}%
\pgfpathlineto{\pgfqpoint{0.955782in}{1.893364in}}%
\pgfusepath{stroke}%
\end{pgfscope}%
\begin{pgfscope}%
\pgfpathrectangle{\pgfqpoint{0.354000in}{0.354000in}}{\pgfqpoint{0.596782in}{2.419000in}} %
\pgfusepath{clip}%
\pgfsetbuttcap%
\pgfsetroundjoin%
\pgfsetlinewidth{1.505625pt}%
\definecolor{currentstroke}{rgb}{0.121569,0.466667,0.705882}%
\pgfsetstrokecolor{currentstroke}%
\pgfsetstrokeopacity{0.100000}%
\pgfsetdash{{2.200000pt}{2.200000pt}}{0.000000pt}%
\pgfpathmoveto{\pgfqpoint{0.354000in}{2.113273in}}%
\pgfpathlineto{\pgfqpoint{0.955782in}{2.113273in}}%
\pgfusepath{stroke}%
\end{pgfscope}%
\begin{pgfscope}%
\pgfpathrectangle{\pgfqpoint{0.354000in}{0.354000in}}{\pgfqpoint{0.596782in}{2.419000in}} %
\pgfusepath{clip}%
\pgfsetbuttcap%
\pgfsetroundjoin%
\pgfsetlinewidth{1.505625pt}%
\definecolor{currentstroke}{rgb}{0.121569,0.466667,0.705882}%
\pgfsetstrokecolor{currentstroke}%
\pgfsetstrokeopacity{0.100000}%
\pgfsetdash{{2.200000pt}{2.200000pt}}{0.000000pt}%
\pgfpathmoveto{\pgfqpoint{0.354000in}{2.333182in}}%
\pgfpathlineto{\pgfqpoint{0.955782in}{2.333182in}}%
\pgfusepath{stroke}%
\end{pgfscope}%
\begin{pgfscope}%
\pgfpathrectangle{\pgfqpoint{0.354000in}{0.354000in}}{\pgfqpoint{0.596782in}{2.419000in}} %
\pgfusepath{clip}%
\pgfsetbuttcap%
\pgfsetroundjoin%
\pgfsetlinewidth{1.505625pt}%
\definecolor{currentstroke}{rgb}{0.839216,0.152941,0.156863}%
\pgfsetstrokecolor{currentstroke}%
\pgfsetstrokeopacity{0.100000}%
\pgfsetdash{{2.200000pt}{2.200000pt}}{0.000000pt}%
\pgfpathmoveto{\pgfqpoint{0.354000in}{2.553091in}}%
\pgfpathlineto{\pgfqpoint{0.955782in}{2.553091in}}%
\pgfusepath{stroke}%
\end{pgfscope}%
\begin{pgfscope}%
\pgfsetrectcap%
\pgfsetmiterjoin%
\pgfsetlinewidth{0.803000pt}%
\definecolor{currentstroke}{rgb}{0.000000,0.000000,0.000000}%
\pgfsetstrokecolor{currentstroke}%
\pgfsetdash{}{0pt}%
\pgfpathmoveto{\pgfqpoint{0.354000in}{0.354000in}}%
\pgfpathlineto{\pgfqpoint{0.354000in}{2.773000in}}%
\pgfusepath{stroke}%
\end{pgfscope}%
\begin{pgfscope}%
\pgfsetrectcap%
\pgfsetmiterjoin%
\pgfsetlinewidth{0.803000pt}%
\definecolor{currentstroke}{rgb}{0.000000,0.000000,0.000000}%
\pgfsetstrokecolor{currentstroke}%
\pgfsetdash{}{0pt}%
\pgfpathmoveto{\pgfqpoint{0.354000in}{0.354000in}}%
\pgfpathlineto{\pgfqpoint{0.950782in}{0.354000in}}%
\pgfusepath{stroke}%
\end{pgfscope}%
\begin{pgfscope}%
\pgfsetrectcap%
\pgfsetmiterjoin%
\pgfsetlinewidth{0.803000pt}%
\definecolor{currentstroke}{rgb}{0.000000,0.000000,0.000000}%
\pgfsetstrokecolor{currentstroke}%
\pgfsetdash{}{0pt}%
\pgfpathmoveto{\pgfqpoint{0.354000in}{2.773000in}}%
\pgfpathlineto{\pgfqpoint{0.950782in}{2.773000in}}%
\pgfusepath{stroke}%
\end{pgfscope}%
\begin{pgfscope}%
\pgfsetbuttcap%
\pgfsetmiterjoin%
\definecolor{currentfill}{rgb}{1.000000,1.000000,1.000000}%
\pgfsetfillcolor{currentfill}%
\pgfsetlinewidth{1.003750pt}%
\definecolor{currentstroke}{rgb}{0.800000,0.800000,0.800000}%
\pgfsetstrokecolor{currentstroke}%
\pgfsetdash{}{0pt}%
\pgfpathmoveto{\pgfqpoint{0.495668in}{0.462774in}}%
\pgfpathlineto{\pgfqpoint{2.323183in}{0.462774in}}%
\pgfpathquadraticcurveto{\pgfqpoint{2.345405in}{0.462774in}}{\pgfqpoint{2.345405in}{0.484996in}}%
\pgfpathlineto{\pgfqpoint{2.345405in}{1.117030in}}%
\pgfpathquadraticcurveto{\pgfqpoint{2.345405in}{1.139252in}}{\pgfqpoint{2.323183in}{1.139252in}}%
\pgfpathlineto{\pgfqpoint{0.495668in}{1.139252in}}%
\pgfpathquadraticcurveto{\pgfqpoint{0.473446in}{1.139252in}}{\pgfqpoint{0.473446in}{1.117030in}}%
\pgfpathlineto{\pgfqpoint{0.473446in}{0.484996in}}%
\pgfpathquadraticcurveto{\pgfqpoint{0.473446in}{0.462774in}}{\pgfqpoint{0.495668in}{0.462774in}}%
\pgfpathclose%
\pgfusepath{stroke,fill}%
\end{pgfscope}%
\begin{pgfscope}%
\pgfsetbuttcap%
\pgfsetmiterjoin%
\definecolor{currentfill}{rgb}{0.172549,0.627451,0.172549}%
\pgfsetfillcolor{currentfill}%
\pgfsetfillopacity{0.200000}%
\pgfsetlinewidth{0.000000pt}%
\definecolor{currentstroke}{rgb}{0.000000,0.000000,0.000000}%
\pgfsetstrokecolor{currentstroke}%
\pgfsetstrokeopacity{0.200000}%
\pgfsetdash{}{0pt}%
\pgfpathmoveto{\pgfqpoint{0.517891in}{1.017030in}}%
\pgfpathlineto{\pgfqpoint{0.740113in}{1.017030in}}%
\pgfpathlineto{\pgfqpoint{0.740113in}{1.094808in}}%
\pgfpathlineto{\pgfqpoint{0.517891in}{1.094808in}}%
\pgfpathclose%
\pgfusepath{fill}%
\end{pgfscope}%
\begin{pgfscope}%
\pgftext[x=0.829002in,y=1.017030in,left,base]{\rmfamily\fontsize{8.000000}{9.600000}\selectfont Prediction window}%
\end{pgfscope}%
\begin{pgfscope}%
\pgfsetrectcap%
\pgfsetroundjoin%
\pgfsetlinewidth{1.505625pt}%
\definecolor{currentstroke}{rgb}{0.121569,0.466667,0.705882}%
\pgfsetstrokecolor{currentstroke}%
\pgfsetdash{}{0pt}%
\pgfpathmoveto{\pgfqpoint{0.517891in}{0.895451in}}%
\pgfpathlineto{\pgfqpoint{0.740113in}{0.895451in}}%
\pgfusepath{stroke}%
\end{pgfscope}%
\begin{pgfscope}%
\pgfsetbuttcap%
\pgfsetmiterjoin%
\definecolor{currentfill}{rgb}{0.121569,0.466667,0.705882}%
\pgfsetfillcolor{currentfill}%
\pgfsetlinewidth{1.003750pt}%
\definecolor{currentstroke}{rgb}{0.121569,0.466667,0.705882}%
\pgfsetstrokecolor{currentstroke}%
\pgfsetdash{}{0pt}%
\pgfsys@defobject{currentmarker}{\pgfqpoint{-0.058926in}{-0.058926in}}{\pgfqpoint{0.058926in}{0.058926in}}{%
\pgfpathmoveto{\pgfqpoint{-0.000000in}{-0.058926in}}%
\pgfpathlineto{\pgfqpoint{0.058926in}{0.000000in}}%
\pgfpathlineto{\pgfqpoint{0.000000in}{0.058926in}}%
\pgfpathlineto{\pgfqpoint{-0.058926in}{0.000000in}}%
\pgfpathclose%
\pgfusepath{stroke,fill}%
}%
\begin{pgfscope}%
\pgfsys@transformshift{0.629002in}{0.895451in}%
\pgfsys@useobject{currentmarker}{}%
\end{pgfscope}%
\end{pgfscope}%
\begin{pgfscope}%
\pgftext[x=0.829002in,y=0.856562in,left,base]{\rmfamily\fontsize{8.000000}{9.600000}\selectfont Returning user (uncensored)}%
\end{pgfscope}%
\begin{pgfscope}%
\pgfsetrectcap%
\pgfsetroundjoin%
\pgfsetlinewidth{1.505625pt}%
\definecolor{currentstroke}{rgb}{0.839216,0.152941,0.156863}%
\pgfsetstrokecolor{currentstroke}%
\pgfsetdash{}{0pt}%
\pgfpathmoveto{\pgfqpoint{0.517891in}{0.728811in}}%
\pgfpathlineto{\pgfqpoint{0.740113in}{0.728811in}}%
\pgfusepath{stroke}%
\end{pgfscope}%
\begin{pgfscope}%
\pgfsetbuttcap%
\pgfsetmiterjoin%
\definecolor{currentfill}{rgb}{0.839216,0.152941,0.156863}%
\pgfsetfillcolor{currentfill}%
\pgfsetlinewidth{1.003750pt}%
\definecolor{currentstroke}{rgb}{0.839216,0.152941,0.156863}%
\pgfsetstrokecolor{currentstroke}%
\pgfsetdash{}{0pt}%
\pgfsys@defobject{currentmarker}{\pgfqpoint{-0.058926in}{-0.058926in}}{\pgfqpoint{0.058926in}{0.058926in}}{%
\pgfpathmoveto{\pgfqpoint{-0.000000in}{-0.058926in}}%
\pgfpathlineto{\pgfqpoint{0.058926in}{0.000000in}}%
\pgfpathlineto{\pgfqpoint{0.000000in}{0.058926in}}%
\pgfpathlineto{\pgfqpoint{-0.058926in}{0.000000in}}%
\pgfpathclose%
\pgfusepath{stroke,fill}%
}%
\begin{pgfscope}%
\pgfsys@transformshift{0.629002in}{0.728811in}%
\pgfsys@useobject{currentmarker}{}%
\end{pgfscope}%
\end{pgfscope}%
\begin{pgfscope}%
\pgftext[x=0.829002in,y=0.689922in,left,base]{\rmfamily\fontsize{8.000000}{9.600000}\selectfont Non-returning user (censored)}%
\end{pgfscope}%
\begin{pgfscope}%
\pgfsetbuttcap%
\pgfsetroundjoin%
\definecolor{currentfill}{rgb}{0.000000,0.000000,0.000000}%
\pgfsetfillcolor{currentfill}%
\pgfsetfillopacity{0.300000}%
\pgfsetlinewidth{1.003750pt}%
\definecolor{currentstroke}{rgb}{0.000000,0.000000,0.000000}%
\pgfsetstrokecolor{currentstroke}%
\pgfsetstrokeopacity{0.300000}%
\pgfsetdash{}{0pt}%
\pgfpathmoveto{\pgfqpoint{0.629002in}{0.499059in}}%
\pgfpathlineto{\pgfqpoint{0.687927in}{0.557984in}}%
\pgfpathlineto{\pgfqpoint{0.629002in}{0.616910in}}%
\pgfpathlineto{\pgfqpoint{0.570076in}{0.557984in}}%
\pgfpathclose%
\pgfusepath{stroke,fill}%
\end{pgfscope}%
\begin{pgfscope}%
\pgftext[x=0.829002in,y=0.528818in,left,base]{\rmfamily\fontsize{8.000000}{9.600000}\selectfont Session}%
\end{pgfscope}%
\end{pgfpicture}%
\makeatother%
\endgroup%

%% file: samplepaper.bbl
\begin{thebibliography}{10}
\providecommand{\url}[1]{\texttt{#1}}
\providecommand{\urlprefix}{URL }
\providecommand{\doi}[1]{https://doi.org/#1}

\bibitem{Bengio:1994:LLD:2325857.2328340}
Bengio, Y., Simard, P., Frasconi, P.: Learning long-term dependencies with
  gradient descent is difficult. IEEE Transactions on Neural Networks
  \textbf{5}(2),  157--166 (1994)

\bibitem{Benson:2016:MUC:2872427.2883024}
Benson, A.R., Kumar, R., Tomkins, A.: Modeling user consumption sequences. In:
  WWW'16. pp. 519--529 (2016)

\bibitem{breslow1974covariance}
Breslow, N.: Covariance analysis of censored survival data. Biometrics
  \textbf{30}(1),  89--99 (1974)

\bibitem{cai2007time}
Cai, X., Zhang, N., Venayagamoorthy, G.K., Wunsch, D.C.: Time series prediction
  with recurrent neural networks trained by a hybrid {PSO--EA} algorithm.
  Neurocomputing  \textbf{70}(13),  2342--2353 (2007)

\bibitem{Chamberlain:2017:CLV:3097983.3098123}
Chamberlain, B.P., Cardoso, A., Liu, C.H.B., Pagliari, R., Deisenroth, M.P.:
  Customer lifetime value prediction using embeddings. In: KDD'17. pp.
  1753--1762. ACM (2017)

\bibitem{chandra2012cooperative}
Chandra, R., Zhang, M.: Cooperative coevolution of elman recurrent neural
  networks for chaotic time series prediction. Neurocomputing  \textbf{86},
  116--123 (2012)

\bibitem{cheng2016wide}
Cheng, H.T., Koc, L., Harmsen, J., Shaked, T., Chandra, T., Aradhye, H.,
  Anderson, G., Corrado, G., Chai, W., Ispir, M., Anil, R., Haque, Z., Hong,
  L., Jain, V., Liu, X., Shah, H.: Wide \& deep learning for recommender
  systems. In: DLRS 2016 (RecSys'16). pp. 7--10. ACM (2016)

\bibitem{cho2014properties}
Cho, K., van Merri{\"e}nboer, B., Bahdanau, D., Bengio, Y.: On the properties
  of neural machine translation: Encoder-decoder approaches. arXiv preprint
  arXiv:1409.1259  (2014)

\bibitem{chollet2015keras}
Chollet, F., et~al.: Keras. \url{https://github.com/fchollet/keras} (2015)

\bibitem{covington45530}
Covington, P., Adams, J., Sargin, E.: Deep neural networks for youtube
  recommendations. In: RecSys'16. pp. 191--198. ACM (2016)

\bibitem{10.2307/2985181}
Cox, D.R.: {Regression Models and Life-Tables}. Journal of the Royal
  Statistical Society. Series B (Methodological)  \textbf{34}(2),  187--220
  (1972)

\bibitem{cox75}
Cox, D.R.: Partial likelihood. Biometrika  \textbf{62}(2),  269--276 (1975)

\bibitem{cox1984analysis}
Cox, D.R., Oakes, D.: Analysis of Survival Data, vol.~21. CRC Press (1984)

\bibitem{Davidson-Pilon2016Lifelines}
Davidson-Pilon, C.: {Lifelines} (2016),
  \url{https://github.com/camdavidsonpilon/lifelines}

\bibitem{Du2016}
Du, N., Dai, H., Trivedi, R., Upadhyay, U., Gomez-Rodriguez, M., Song, L.:
  Recurrent marked temporal point processes: Embedding event history to vector.
  In: KDD'16. pp. 1555--1564. ACM (2016)

\bibitem{Du:2015:TRR:2969442.2969629}
Du, N., Wang, Y., He, N., Song, L.: Time-sensitive recommendation from
  recurrent user activities. In: NIPS'15. pp. 3492--3500. MIT Press (2015)

\bibitem{efron1977efficiency}
Efron, B.: The efficiency of cox's likelihood function for censored data.
  Journal of the American Statistical Association  \textbf{72}(359),  557--565
  (1977)

\bibitem{flunkert2017deepar}
Flunkert, V., Salinas, D., Gasthaus, J.: Deepar: Probabilistic forecasting with
  autoregressive recurrent networks. arXiv preprint arXiv:1704.04110  (2017)

\bibitem{graves2009novel}
Graves, A., Liwicki, M., Fern{\'a}ndez, S., Bertolami, R., Bunke, H.,
  Schmidhuber, J.: A novel connectionist system for unconstrained handwriting
  recognition. IEEE Transactions on Pattern Analysis and Machine Intelligence
  \textbf{31}(5),  855--868 (2009)

\bibitem{han2004prediction}
Han, M., Xi, J., Xu, S., Yin, F.L.: Prediction of chaotic time series based on
  the recurrent predictor neural network. IEEE Transactions on Signal
  Processing  \textbf{52}(12),  3409--3416 (2004)

\bibitem{Harrell1996MultivariableErrors}
Harrell, F.E., Lee, K.L., Mark, D.B.: Multivariable prognostic models: Issues
  in developing models, evaluating assumptions and adequacy, and measuring and
  reducing errors. Statistics in Medicine  \textbf{15},  361--387 (1996)

\bibitem{10.2307/2334319}
Hawkes, A.G.: {Spectra of Some Self-Exciting and Mutually Exciting Point
  Processes}. Biometrika  \textbf{58}(1),  83--90 (1971)

\bibitem{doi:10.1162/neco.1997.9.8.1735}
Hochreiter, S., Schmidhuber, J.: Long short-term memory. Neural Computation
  \textbf{9}(8),  1735--1780 (1997)

\bibitem{ishwaran2008}
Ishwaran, H., Kogalur, U.B., Blackstone, E.H., Lauer, M.S.: Random survival
  forests. Ann. Appl. Stat.  \textbf{2}(3),  841--860 (2008)

\bibitem{10.2307/2334538}
Kalbfleisch, J.D., Prentice, R.L.: Marginal likelihoods based on cox's
  regression and life model. Biometrika  \textbf{60}(2),  267--278 (1973)

\bibitem{Kapoor:2014:HBA:2623330.2623348}
Kapoor, K., Sun, M., Srivastava, J., Ye, T.: A hazard based approach to user
  return time prediction. In: KDD'14. pp. 1719--1728. ACM (2014)

\bibitem{klein2005survival}
Klein, J.P., Moeschberger, M.L.: Survival analysis: techniques for censored and
  truncated data. Springer Science \& Business Media (2005)

\bibitem{Li:2017:NNC:3041021.3054200}
Li, L., Jing, H., Tong, H., Yang, J., He, Q., Chen, B.C.: {NEMO}: Next career
  move prediction with contextual embedding. In: WWW '17 Companion. pp.
  505--513 (2017)

\bibitem{Manzoor:2017:RTT:3097983.3098104}
Manzoor, E., Akoglu, L.: Rush!: Targeted time-limited coupons via purchase
  forecasts. In: KDD'17. pp. 1923--1931. ACM (2017)

\bibitem{mikolov2013efficient}
Mikolov, T., Chen, K., Corrado, G., Dean, J.: Efficient estimation of word
  representations in vector space. arXiv preprint arXiv:1301.3781  (2013)

\bibitem{deepSurvivalAnalysis}
Rajesh, R., Perotte, A., Elhadad, N., Blei, D.: {Deep Survival Analysis}. In:
  Proceedings of the 1st Machine Learning for Healthcare Conference. pp.
  101--114 (2016)

\bibitem{Rodriguez2010SurvivalModels}
Rodr{\'{i}}guez, G.: {Survival Models}. In: Course Notes for Generalized Linear
  Statistical Models (2010),
  \url{http://data.princeton.edu/wws509/notes/c7.pdf}

\bibitem{sutskever2014sequence}
Sutskever, I., Vinyals, O., Le, Q.V.: Sequence to sequence learning with neural
  networks. In: NIPS'14. pp. 3104--3112 (2014)

\bibitem{vinyals2015show}
Vinyals, O., Toshev, A., Bengio, S., Erhan, D.: Show and tell: A neural image
  caption generator. In: CVPR'15. pp. 3156--3164 (2015)

\bibitem{1604.05377}
Wangperawong, A., Brun, C., Laudy, O., Pavasuthipaisit, R.: Churn analysis
  using deep convolutional neural networks and autoencoders. arXiv preprint
  arXiv:1604.05377  (2016)

\end{thebibliography}
